\AddToHook{package/hyperref/before}{\RequirePackage{titlesec}}
\documentclass{gensi}

\usepackage{amsmath,amsfonts,bm}

\def\eqref#1{equation~\ref{#1}}

\def\1{\bm{1}}

\DeclareMathAlphabet{\mathsfit}{\encodingdefault}{\sfdefault}{m}{sl}
\SetMathAlphabet{\mathsfit}{bold}{\encodingdefault}{\sfdefault}{bx}{n}

\usepackage{amssymb}
\usepackage{enumitem}
\usepackage{colortbl}
\usepackage{fix-cm}
\usepackage{trion-layout}

\title{Trion-1: Unifying Language, Protein Sequence, and Structure with Diffusion Language Models}

\affiliation[1]{Generative Symbolic Intelligence Lab (GenSI), Tsinghua University}
\affiliation[2]{Institute for AI Industry Research (AIR), Tsinghua University}
\affiliation[3]{Shanghai Artificial Intelligence Laboratory}

\abstract{
We envision a general-purpose protein foundation model as a unified, interactive model of sequence,
structure, and function, capable of understanding, generation, and cross-modal translation while
retaining the instruction-following and reasoning capabilities of modern LLMs. 
Toward this vision, we introduce \textbf{Trion-1}, the first native, unified, multimodal diffusion
language model (dLLM) for proteins. 
Trion-1 is built on two key design choices: (1) a unified discrete token space that jointly
represents sequence, structure, and natural language; (2) a dLLM backbone that replaces strict
left-to-right factorization with bidirectional or semi-bidirectional token interactions, enabling
global residue communication and long-range contact modeling that are naturally suited to structure
modeling. 
This formulation unifies six core capabilities: folding, inverse folding, sequence understanding,
structure understanding, sequence design, and structure design, while retaining the native language
capabilities of general-purpose LLMs. On established benchmarks, Trion-1 achieves broadly
competitive performance. In folding, Trion-1 achieves a TM-score above 0.8 with a single sample
under strict homology control, matching or exceeding the existing tri-modal model using 200
samples. For sequence understanding, Trion-1 performs on par with top-tier LLMs such as
Claude Opus 4.7 (62.37\%). 
These results support well-aligned sequence--structure--language representations and strong
cross-modal generative capabilities. 
}

\begin{document}
\maketitle

\par\medskip
\noindent\begin{minipage}{\textwidth}
  \centering
  \includegraphics[width=\linewidth,height=0.36\textheight,keepaspectratio]{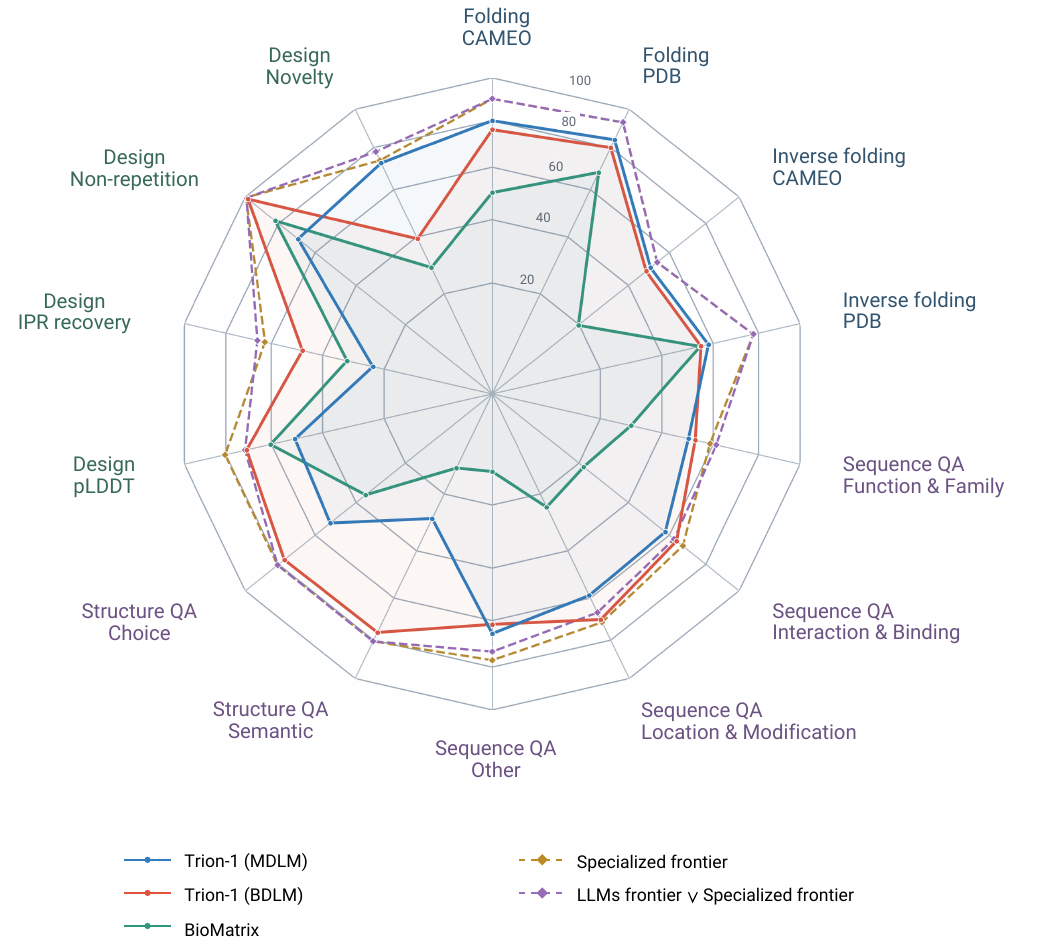}
  \captionof{figure}{Performance Rader.}
  \label{fig:performance-radar}
\end{minipage}
\clearpage

\FloatBarrier
\section{Introduction}
\label{sec:intro}

Proteins constitute the molecular machinery of life, carrying out the catalytic, structural,
signaling, and regulatory processes that sustain cells and organisms. A comprehensive understanding
of proteins draws on three complementary modalities: (1) \emph{amino acid sequences} describe
molecular identity and evolutionary constraints \citep{ESM2}. (2) \emph{3D structures} represent the
physical and geometric constraints that govern molecular interactions, thereby revealing the
mechanistic basis of protein function \citep{Structure}. (3) \emph{function descriptions} capture
not only specific protein functions but also the broader body of human knowledge accumulated through
biological research \citep{gene2021, uniprot2023}. 

Importantly, these modalities are not independent views of proteins. We distill protein modeling
into six meta-capabilities spanning the directed transformations among sequence, structure, and
function: \textbf{folding}, \textbf{inverse folding}, \textbf{sequence understanding},
\textbf{structure understanding}, \textbf{sequence design}, and \textbf{structure design}. Rather
than isolated tasks, these capabilities serve as composable primitives whose combinations give rise
to diverse and increasingly complex workflows.

Substantial progress has been made through specialized protein models; in
Sec.~\ref{sec:specialized-model}, we review these developments and revisit several long-overlooked connections
to modern language modeling. Nevertheless, protein modeling remains fragmented across specialized
systems and largely disconnected from general-purpose LLMs. We identify two key requirements for
next-generation general-purpose protein foundation models:
\begin{itemize}[leftmargin=20pt]
\item \textbf{Unified protein framework.} A single model should enable knowledge sharing and mutual
reinforcement across tasks, without heavily relying on cross-model interfaces or task-specific adaptation.
\item \textbf{Native language capabilities.} The model should retain the instruction-following,
reasoning, and interactive capabilities native to general-purpose LLMs.
\end{itemize}

Early attempts explored unifying heterogeneous protein modalities within autoregressive frameworks
\citep{ProtTex, HelixProtX}, roughly in parallel with the emergence of unified vision models
\citep{Chameleon, Emu3}. Yet these approaches remained limited on structure-intensive tasks such as
protein folding. Theoretically, structural contacts do not follow a left-to-right sequence order,
so autoregressive generation may commit to early predictions with incomplete context and accumulate
errors. Experimentally, existing tri-modal autoregressive models show limited folding performance
under single-sample evaluation, leaving open \textit{whether the autoregressive paradigm itself provides the right
foundation for unified protein modeling}.

We introduce \textbf{Trion-1}, to our knowledge the first native, unified, tri-modal diffusion
language model (dLLM) for proteins. Trion-1 builds on two key design choices: (1) a shared
discrete token space that jointly represents protein sequence, structure, and natural language; and
(2) a non-causal diffusion backbone that relaxes strict left-to-right factorization, enabling global
residue interactions modeling within an LLM architecture. Together, these designs unify the six
directed capabilities across the sequence--structure--function space within a single generative
model. Across established benchmarks, Trion-1 achieves broadly competitive performance, with
particularly substantial improvements over existing unified tri-modal models on protein folding.
 
We highlight our contributions as follows:
\begin{itemize}[leftmargin=20pt]

\item \textbf{Framework.}
We develop Trion-1, a native unified protein dLLM that brings sequence, structure, and natural
language into a shared discrete token space and models them with a non-causal diffusion backbone.
Guided by the intrinsic compatibility between specialized protein modeling and modern language-model
primitives, Trion-1 unifies six core capabilities within a single generative framework while
retaining native language interaction, and further supports their interleaved composition.

\item \textbf{Capability.}
At the 4B scale, Trion-1 achieves competitive performance across all six core capabilities and
substantially strengthens unified tri-modal folding, reaching TM$_{\mathrm{bb}}$ 0.84 with a single
unified MDLM checkpoint---comparable to prior tri-modal models evaluated with a 200$\times$ larger
inference budget. On GeneralQA, Trion-1 further performs on par with frontier models such as
Claude Opus 4.7 \citep{Claude47} and AMix2 \citep{Amix2}.

\end{itemize}

\FloatBarrier
\section{Related Work}
\label{sec:related}

\subsection{Diffusion Language Models}
Autoregressive (AR) language models generate sequences through a left-to-right factorization, which
imposes a fixed generation order. Diffusion language models (dLLMs) provide an alternative by
modeling text generation as an iterative denoising process, allowing tokens to be predicted with
bidirectional context. Recent advances have enabled dLLMs to rapidly close the performance gap with
AR LLMs \citep{Llada2.2, DiffusionGemma}, establishing diffusion-based language modeling as a
promising alternative to the dominant paradigm.

Foundational studies \citep{MDLM, MD4, RADD} established the formulation of objective as \emph{a
time-independent, weighted sum of standard cross-entropy losses}. Recent studies have further
explored block-based formulations that introduce semi-autoregressive generation \citep{BD3LM}, along
with scalable training and sampling strategies \citep{dllm, Llada2.0}, and novel architectures
\citep{Flare, DiffusionGemma}.

\subsection{Specialized Model}
\label{sec:specialized-model}
We review recent progress in specialized protein models and the key inductive biases behind their
success, while highlighting their intrinsic connections to the modeling capabilities of modern LLMs.

\textbf{Folding.}
Protein folding maps amino acid sequences to 3D structures. Modern folding systems rely heavily on
triangular residue interactions to capture higher-order geometric consistency and long-range
contacts \citep{AF2, AF3}, with recent advances driven by scaling \citep{SeedFold}, distillation
\citep{Protenix-mini, Protenix-mini+}, and richer conditioning \citep{Boltz1, Boltz2, Chai1, Chai2}.
From a graph perspective, the effectiveness of triangular modeling can be understood through two
complementary components: \emph{translation self-loops}, which map residue context into local
structure, and \emph{interaction edges}, which propagate information across residues to enforce
global consistency. Although LLMs lack explicit interaction-edge modules, recent results show that
strong folding can emerge from largely data-driven Transformer architectures \citep{SimpleFold}.

\textbf{Inverse folding.}
Modern inverse folding models \citep{PMPNN, LigandMPNN, ABACUS} typically encode protein backbones
into geometry-aware representations before light-weight sequence decoding. This closely mirrors the
encoder of a protein structure tokenizer, which likewise maps 3D backbone geometry into a compact
latent representation, suggesting a natural bridge from inverse folding to language-model-based
generation.

\textbf{Sequence/Structure understanding.}
Protein understanding is shifting from fixed-label prediction toward open-ended functional
semantics. Sequence models increasingly align proteins with textual description \citep{ProtT5,
ProtST}, while structure-aware models inject geometric context \citep{GVP, GearNet}. This makes
natural language a natural interface between molecular representations and biological knowledge.

\textbf{Sequence/Structure design.}
Protein design is likewise moving toward richer conditioning: sequence models progress from
functional tags \citep{Progen} to language descriptions \citep{ProteinDT, ProtDAT, Amix2} and motif
constraints \citep{EvoDiff, DPLM}, while structure generators incorporate increasingly detailed
physicochemical and geometric constraints \citep{RFDiffusion, RFDiffusion2, Proteina, LaProteina}.
This trend naturally aligns protein design with instruction-conditioned generation in modern LLMs,
where heterogeneous design intents can be expressed and composed through a shared interface.

\subsection{Unified Tri-modal Models}
Early tri-modal unified models primarily focus on representation-level alignment, mapping sequence,
structure, and text into a shared embedding space through contrastive learning \citep{ProTrek,
ProteinAligner}. ESM3 \citep{ESM3} moves beyond alignment toward generative unification. BioMatrix
explores native tri-modal generation with a single AR LLM backbone \citep{Biomatrix}, while our work
concurrently investigates the analogous unified formulation with dLLM framework.

\FloatBarrier
\section{Methods}

\begin{figure}[!t]
  \centering
  \includegraphics[width=1.0\linewidth]{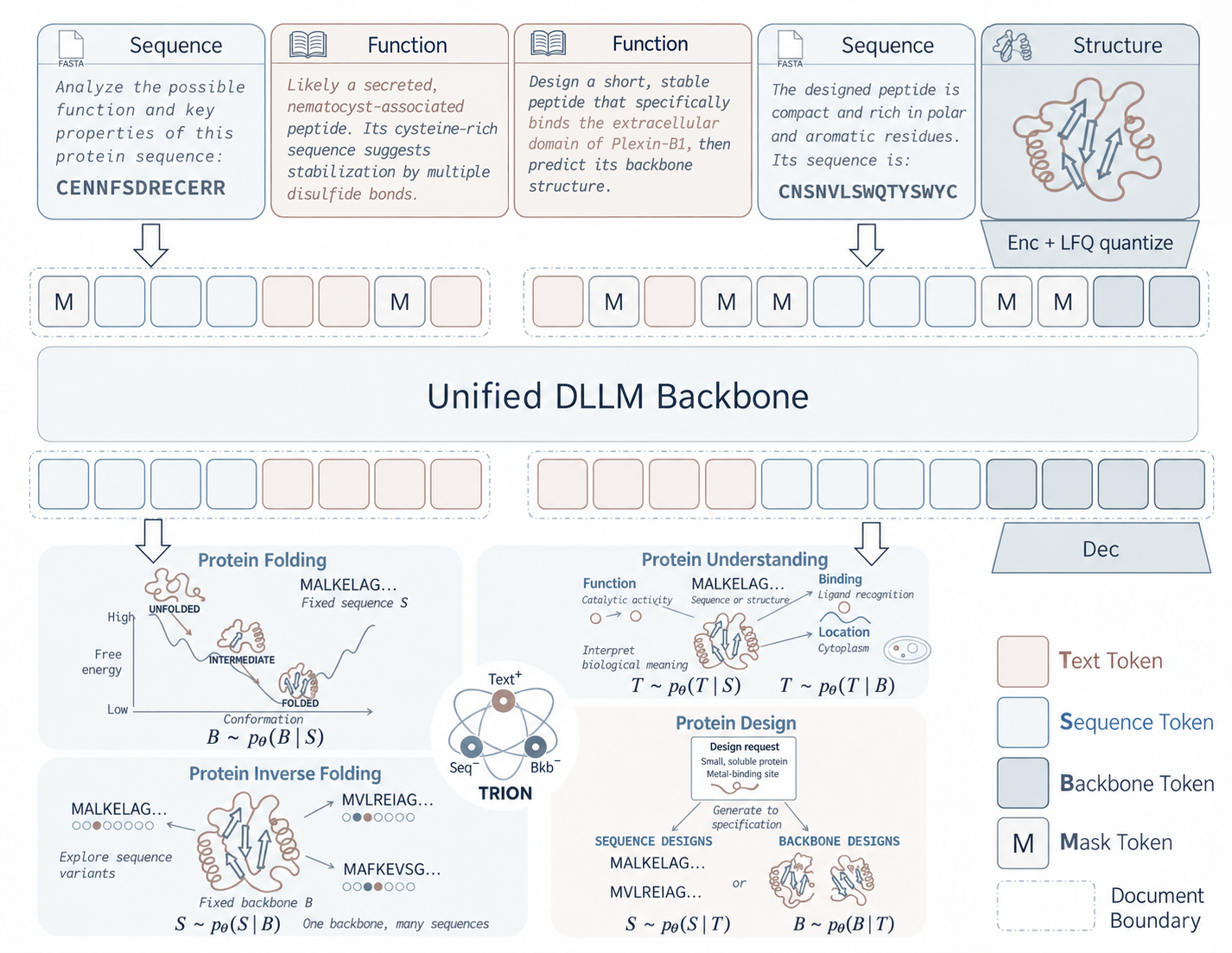}
  \caption{Overview of Trion-1.}
  \label{fig:overview}
\end{figure}

\subsection{Shared Embedding}
\label{sec:m1}

We adopt token-space early-fusion. Specifically, we extend the Qwen3 vocabulary \citep{Qwen3} with
dedicated sequence and quantized structure tokens, allowing all modalities to be processed by a
single dLLM from the first layer without separate modality encoders or cross-attention modules. A
key challenge is how to discretize continuous protein structures into an LLM-compatible token space.
We favor a tokenizer with three properties:

\begin{itemize}[leftmargin=20pt]
\item \textbf{Generative.}
The tokenizer should preserve sufficient geometric information for high-fidelity backbone
reconstruction, rather than serving only retrieval search \citep{FoldSeek}. This is naturally
supported by autoencoding objectives \citep{VAE, VQVAE}, optionally regularized by auxiliary
semantic supervision \citep{ESM3}. Recent work further shows that diffusion- or flow-matching-based
decoders can substantially improve reconstruction fidelity, both in visual tokenization
\citep{FlowMo, DiTo} and protein structure tokenization \citep{Kanzi, Yeti}, by modeling a
conditional distribution over reconstructions rather than relying on a deterministic decoder.

\item \textbf{Local.}
Each token should correspond to a local structural context, preserving residue-level positional
correspondence and allowing contiguous tokens to represent local substructures and motifs. This
contrasts with global tokenization schemes \citep{Adaptive}, which may weaken residue-level
grounding.

\item \textbf{Codebook-free.}
Codebook-free quantization, such as LFQ \citep{LFQ}, maps tokens to vertices of an $n$-dimensional
hypercube, providing $2^n$ discrete codes with a natural Hamming geometry that can be readily
absorbed into the extended embedding space, as shown in \ref{sec:e2.1}.
\end{itemize}

Codebook-free quantization is a convenient choice rather than a strict requirement; more generally,
the tokenizer geometry should be faithfully represented in the language-model embedding space, which
can also emerge through large-scale training \citep{Emu3, Chameleon}. Following these principles, we
use the DPLM2 structure tokenizer \citep{DPLM2}.

\subsection{Unified Architecture}
Given the unified token space, we use a single dLLM backbone to jointly model protein sequence,
backbone structure, and natural language, as illustrated in \ref{fig:overview}. All modalities share
the same Transformer and can flexibly serve as conditions or generation targets, enabling both
intra- and cross-modal generation without modality-specific heads. 

We instantiate two variants with the same architecture but different generation factorizations:
Trion-1 (MDLM) uses fully non-causal masked diffusion, while Trion-1 (BDLM) applies
bidirectional diffusion within blocks and causal factorization across blocks. This provides a
controlled comparison of how bidirectionality affects protein modeling. 

{
\setlength{\abovedisplayskip}{2pt}
\setlength{\belowdisplayskip}{2pt}
\begin{equation}
\begin{aligned}
\mathcal{L}_{\text{MDLM}}(\theta)
&=
-\mathbb{E}_{t, x_0 \sim p_{\text{data}}}
\mathbb{E}_{x_t \sim q(x_t \mid x_0)}
\left[
\frac{1}{t}
\sum_{i=1}^{L}
\mathbf{1}\!\left[x_t^{i}=\mathbf{M}\right]
\log p_\theta\!\left(x_0^{i}\mid x_t\right)
\right]
\\
\mathcal{L}_{\text{BDLM}}(\theta)
&=
-\mathbb{E}_{t, x_0 \sim p_{\text{data}}}
\mathbb{E}_{x_t \sim q(x_t \mid x_0)}
\left[
\frac{1}{t}
\sum_{b=1}^{B}
\sum_{i \in \mathcal{I}_b}
\mathbf{1}[x_t^i=\mathbf{M}]
\log
p_\theta
\left(
x_0^i
\mid
x_0^{[<b]},\,x_t^{[b]}
\right)
\right],
\end{aligned}
\end{equation}
}

Here, $\mathbf{M}$ denotes the mask token, $L$ is the sequence length, and $B$ is the number of
blocks with $\mathcal{I}_b$ denoting the token indices in block $b$. For BDLM, $x_0^{[<b]}$ denotes
the clean tokens in preceding blocks, while $x_t^{[b]}$ denotes the corrupted tokens in the current
block; thus diffusion is bidirectional within each block but causal across blocks.

\subsection{Multi-stage Training}
Protein sequence bridges language and backbone structure in our training strategy. We first align
sequence with language during continual pre-training (CPT), then connect both to structure through
multimodal supervised fine-tuning (SFT). Given the scarcity of structural data, structure is
introduced only during SFT. Starting from Qwen3-4B, we proceed through AR-CPT for protein knowledge
acquisition, A2D-CPT for adaptation to diffusion modeling~\citep{dllm}, and SFT for all six directed
cross-modal transformations (Table~\ref{tab:training_stages}).

\begin{table}[!ht]
\centering
\small
\caption{Training configuration. T/S/B: text/sequence/backbone; ST: sequence--text pairs; X2X: all
six cross-modal directions. Lengths are in tokens.}
\label{tab:training_stages}
\begingroup
\setlength{\tabcolsep}{5pt}
\renewcommand{\arraystretch}{1.12}
\begin{tabular}{@{}lccc@{}}
\toprule
& \textbf{Stage 1} & \textbf{Stage 2} & \textbf{Stage 3} \\
\midrule
Training Paradigm & AR-CPT & A2D-CPT & SFT \\
Context Length & 4096 & 4096 & 4096 \\
Block Size & 1 & $8 \rightarrow 16 \rightarrow 32$ & 32 (bdlm) / 4096 (mdlm) \\
Assistant Mask & $\times$ & $\times$ & $\checkmark$ \\
Document Mask & $\times$ & $\times$ & $\checkmark$ \\
Task Type & T + S + ST & T + S + ST & X2X \\
\bottomrule
\end{tabular}
\endgroup
\end{table}


\FloatBarrier
\section{Experiments}
\subsection{Quantitative Performance}

\subsubsection{Evaluation Settings}
\textbf{Benchmarks}. We prioritize evaluations that reflect the current frontier of protein
modeling. For example, ProteinArena \citep{Amix2} refines PDFBench \citep{PDFBench}, providing a
more recent and rigorous benchmark for sequence understanding and design. The benchmarks are
summarized as follows:
\begin{itemize}[leftmargin=20pt]
\item \textbf{Folding Tasks}: CAMEO 2022 \citep{Eigenfold}, PDB-date \citep{MultiFlow}
\item \textbf{Inverse Folding Tasks}: CAMEO 2022 \citep{Eigenfold}, PDB-date \citep{MultiFlow}
\item \textbf{Protein Analysis Tasks}: ProteinArena \citep{Amix2}, OPI-struct \cite{Stella} 
\item \textbf{Protein Design Tasks}: ProteinArena \citep{Amix2}
\end{itemize}
\textbf{Baselines}.
Our baseline selection follows two principles. \emph{Capability}: we include specialized protein
models and frontier LLMs as task-level performance references. To maintain comparable input
requirements, we exclude AlphaFold-family configurations that require MSA inputs. \emph{Paradigm}:
we include native multimodal models spanning text--sequence, sequence--structure, and
text--sequence--structure modeling, covering autoregressive, masked diffusion, and block diffusion
generation paradigms. Together, these comparisons assess both Trion-1's task performance and the
implications of its modality coverage and generation paradigm.

\FloatBarrier
\subsubsection{Folding and Inverse folding}
\label{sec:e1.1}
As shown in Table~\ref{tab:folding-inverse-folding-results}, Trion-1 substantially pushes
forward the folding capability of unified protein models, while remaining competitive on inverse
folding. We highlight two observations from these results.

\suppressfloats[t]
\begin{table}[htbp]
    \centering
    \newcommand{\foldTMpair}[2]{\makebox[2.2em][r]{#1}\,/\,\makebox[2.2em][l]{#2}}
    \newcommand{\foldOracle}{\cellcolor{gray!15}}
    \caption{\textbf{Folding and inverse folding.} Means are reported.
    Folding: TMaa over common atoms, TMus from C$\alpha$ USalign, and RMSDaa.
    Yellow: our models. Gray: per-metric group best (ties included).
    Local BioMatrix uses the public merged checkpoint.
    @F/@I: folding/inverse-folding sample counts; @200 uses metric-wise oracle selection,
    while @50 selects the highest scTMbb and reports that same candidate's AAR.
    @F/@I$^\dagger$: paper-reported counts (length-filtered)~\citep{Biomatrix}.
    $*$: DPLM-2 paper. Cited results retain original metric protocols; --: unavailable.}
    \label{tab:folding-inverse-folding-results}
    \footnotesize
    \setlength{\tabcolsep}{3pt}
    \renewcommand{\arraystretch}{1.08}
    \FitTableToPage{%
    \begin{tabular}{@{}l@{\hspace{3pt}}c@{\hspace{3pt}}*{8}{c}@{}}
        \toprule
        \multirow{3}{*}{\textbf{Method}}
            & \multirow{3}{*}{\textbf{Params}}
            & \multicolumn{4}{c}{\textbf{Folding}}
            & \multicolumn{4}{c}{\textbf{Inverse Folding}} \\
        \cmidrule[0.3pt](lr){3-6}\cmidrule[0.3pt](lr){7-10}
            & & \multicolumn{2}{c}{CAMEO 2022 (183)}
            & \multicolumn{2}{c}{PDB date split (449)}
            & \multicolumn{2}{c}{CAMEO 2022 (183)}
            & \multicolumn{2}{c}{PDB date split (449)} \\
        \cmidrule[0.3pt](lr){3-4}\cmidrule[0.3pt](lr){5-6}
        \cmidrule[0.3pt](lr){7-8}\cmidrule[0.3pt](lr){9-10}
            & & \texttt{TMaa/us} $\uparrow$
            & \texttt{RMSDaa} $\downarrow$
            & \texttt{TMaa/us} $\uparrow$
            & \texttt{RMSDaa} $\downarrow$
            & \texttt{AAR} $\uparrow$
            & \texttt{scTMbb} $\uparrow$
            & \texttt{AAR} $\uparrow$
            & \texttt{scTMbb} $\uparrow$ \\
        \midrule
        \multicolumn{10}{l}{\textit{Frontier Folding/Inverse Folding Models}} \\
        \midrule
        \quad ESMFold/IF & 3.7B/0.1B
            & \foldOracle\foldTMpair{0.91}{0.85} 
            & 4.61 & \foldTMpair{0.94}{\colorbox{gray!15}{0.91}} 
            & 3.22 & \foldOracle 52.96 
            & \foldOracle 0.90 & \foldOracle 59.20 & \foldOracle 0.95 \\
        \quad SimpleFold & 3B
            & \foldTMpair{0.90}{0.83} 
            & 4.88 & \foldOracle\foldTMpair{0.95}{0.91} 
            & \foldOracle2.99 & -- 
            & -- & -- & -- \\
        \quad PVQD$^*$ & -
            & \foldTMpair{--}{0.81} & \foldOracle 4.08
            & -- & --
            & -- & -- & -- & -- \\
        \midrule
        \multicolumn{10}{l}{\textit{Sequence--Structure Co-generation Models}} \\
        \midrule
        \quad MultiFlow$^*$ & 21.8M
            & \foldTMpair{0.50}{--} & 17.84
            & \foldTMpair{0.53}{--} & 15.64
            & 32.28 & 0.87 & 37.74 & 0.94 \\
        \quad ESM3 & 1.4B
            & \foldOracle\foldTMpair{0.86}{0.79} 
            & \foldOracle5.46 & \foldTMpair{0.90}{0.84} 
            & 4.10 & 48.53 
            & \foldOracle 0.90 & 51.64 & \foldOracle 0.95 \\
        \quad DPLM2 & 3B
            & \foldTMpair{0.83}{0.72} 
            & 7.10 & \foldTMpair{0.90}{0.81} 
            & 4.28 & 51.48 
            & 0.88 & 61.18 & 0.92 \\
        \quad DPLM2 & 0.7B
            & \foldTMpair{0.80}{0.69} 
            & 7.81 & \foldTMpair{0.87}{0.77} 
            & 5.28 & 48.49 
            & 0.87 & 55.26 & 0.91 \\
        \quad DPLM2.1 & 0.7B
            & \foldTMpair{0.84}{0.74} 
            & 6.61 & \foldOracle\foldTMpair{0.93}{0.88} 
            & \foldOracle2.95 & \foldOracle 53.79 
            & 0.88 & \foldOracle 77.62 & 0.93 \\
        \midrule
        \multicolumn{10}{l}{\textit{Sequence--Structure--Text LLMs}} \\
        \midrule
        \quad BioMatrix@1/@1 & 4B
            & \foldTMpair{0.45}{0.35} 
            & 18.71 & \foldTMpair{0.60}{0.52} 
            & 12.62 & 19.76 
            & 0.57 & 54.67 & 0.75 \\
        \quad BioMatrix@200/@50 & 4B
            & \foldTMpair{0.67}{0.57} 
            & 11.68 & \foldTMpair{0.82}{\colorbox{gray!15}{0.77}} 
            & 5.68 & 21.51 
            & 0.76 & 60.43 & 0.89 \\
        \quad BioMatrix@200/@50$^\dagger$ & 4B
            & -- & --
            & \foldTMpair{0.84}{--} & \foldOracle 5.38
            & -- & -- & \foldOracle 75.50 & 0.86 \\
        \rowcolor{yellow!15}
        \quad Trion-1 (MDLM) & 4B
            & \foldOracle\foldTMpair{0.81}{0.71} 
            & \foldOracle7.83 & \foldOracle\foldTMpair{0.86}{0.77} 
            & 5.64 & \foldOracle 50.48 
            & \foldOracle 0.88 & 58.03 & \foldOracle 0.92 \\
        \rowcolor{yellow!15}
        \quad Trion-1 (BDLM) & 4B
            & \foldTMpair{0.78}{0.66} 
            & 9.16 & \foldTMpair{0.82}{0.73} 
            & 7.28 & 48.32 
            & 0.87 & 55.00 & 0.91 \\
        \bottomrule
    \end{tabular}%
    }
\end{table}

\Needspace{4\baselineskip}
\textbf{Bidirectional context benefits structure modeling.}
On the PDB date split, Trion-1 (MDLM) outperforms BioMatrix@1 in folding,
improving TMaa from 0.60 to 0.86 and reducing RMSDaa from 12.62 to 5.64.
It also reaches a TMaa of 0.86 versus 0.82 for BioMatrix's metric-wise
best-of-200 sampling, with RMSDaa of 5.64 versus 5.68.
Unlike autoregressive generation, protein geometry is not governed
by a left-to-right dependency order: spatially adjacent residues
may be far apart in sequence. Causal attention prevents a residue
from directly accessing such partners when they occur later in
the sequence, whereas bidirectional attention permits this context.
The consistent MDLM-over-BDLM-over-AR advantage supports this interpretation,
which we further investigate in Sect.~\ref{sec:e2.2}.

\textbf{Global geometric consistency remains a challenge.}
Trion-1 (MDLM) reaches a PDB TM$_{\mathrm{aa}}$ of 0.86 without
explicit triangular modules, but still trails ESMFold (0.94).
This gap suggests room to improve how the model coordinates
interactions among residues beyond sequence--structure translation.
We aim to improve this capability through self-training on
model-generated data, preserving the simplicity of modern LLM
architectures rather than introducing computationally intensive
triangular modules.

\FloatBarrier
\subsubsection{Protein Understanding}
\label{sec:e1.2}

We evaluate sequence understanding on GeneralQA and structure understanding on StructureMCQA, the
multiple-choice subset of OPI-Struc. Table~\ref{tab:sequence-structure-understanding-results}
summarizes both tasks.

\suppressfloats[t]
\begin{table}[htbp]
    \centering
    \newcommand{\qaOracle}{\cellcolor{gray!15}}
    \caption{\textbf{Sequence and structure understanding.} Accuracy (\%, $\uparrow$). F\&F:
    Function \& Family; I\&B: Interaction \& Binding; L\&M: Location \& Modification; Other:
    physicochemical/structural properties. Choice: option accuracy; Semantic: final-answer semantic
    correctness. Yellow: our models. Gray: per-metric group best (ties included); --: unavailable. $\dagger$: BioMatrix uses sequence+structure
    input and nearest-option mapping for Choice; its Semantic score uses a different judge prompt.}
    \label{tab:sequence-structure-understanding-results}
    \footnotesize
    \setlength{\tabcolsep}{3pt}
    \renewcommand{\arraystretch}{1.02}
    \begin{tabular}{@{}l@{\hspace{3pt}}c@{\hspace{3pt}}*{7}{c}@{}}
        \toprule
        \multirow{3}{*}{\textbf{Method}}
            & \multirow{3}{*}{\textbf{Params}}
            & \multicolumn{5}{c}{\textbf{GeneralQA}}
            & \multicolumn{2}{c}{\textbf{StructureMCQA}} \\
        \cmidrule[0.3pt](lr){3-7}\cmidrule[0.3pt](lr){8-9}
            & & Overall & F\&F & I\&B & L\&M & Other & Choice & Semantic \\
            & & (481) & (150) & (152) & (153) & (26) & (3,918) & (3,918) \\
        \midrule
        \multicolumn{9}{l}{\textit{Frontier LLMs}} \\
        \midrule
        \quad Claude Opus 4.7         & -- & \qaOracle 62.37 & \qaOracle 61.33 & \qaOracle 62.50 & \qaOracle 66.67 & 42.31 & -- & -- \\
        \quad Claude Opus 4.6         & -- & 48.02 & 32.67 & 48.68 & 58.17 & \qaOracle 73.08 & -- & -- \\
        \quad Gemini 3 Flash Thinking & -- & 49.48 & 42.00 & 46.05 & 59.48 & 53.85 & -- & -- \\
        \quad GPT-5.5                  & -- & 48.44 & 32.00 & 46.05 & 66.01 & 53.85 & -- & -- \\
        \quad DeepSeek-V4-Pro          & 1.6T & 46.57 & 36.67 & 45.39 & 56.21 & 53.85 & -- & -- \\
        \quad Qwen3.7-Max              & -- & 43.45 & 31.33 & 43.42 & 54.25 & 50.00 & -- & -- \\
        \quad GPT-OSS-120B             & 120B & 43.24 & 32.00 & 38.82 & 56.21 & 57.69 & -- & -- \\
        \quad GLM-5.1                  & 0.8T & 38.05 & 26.67 & 37.50 & 49.02 & 42.31 & -- & -- \\
        \midrule
        \multicolumn{9}{l}{\textit{Sequence--Text LLMs}} \\
        \midrule
        \quad AMix2-dLLM & 4B & \qaOracle 66.11 & \qaOracle 58.67 & \qaOracle 67.11 & \qaOracle 71.24 & 73.08 & -- & -- \\
        \quad AMix2-AR   & 4B & 62.58 & 56.00 & 61.84 & 67.32 & \qaOracle 76.92 & -- & -- \\
        \midrule
        \multicolumn{9}{l}{\textit{Sequence--Structure--Text LLMs}} \\
        \midrule
        \quad BioMatrix & 4B & 24.32 & 29.33 & 21.71 & 24.18 & 11.54 & 35.76$^\dagger$ & 12.61$^\dagger$ \\
        \rowcolor{yellow!15}
        \quad Trion-1 (MDLM) & 4B & 56.13 & 50.00 & 57.89 & 58.82 & \qaOracle 65.38 & 52.25 & 28.08 \\
        \rowcolor{yellow!15}
        \quad Trion-1 (BDLM) & 4B & \qaOracle 62.16 & \qaOracle 52.67 & \qaOracle 63.82 & \qaOracle 69.93 & 61.54 & \qaOracle 76.72 & \qaOracle 76.19 \\
        \bottomrule
    \end{tabular}
\end{table}

\textbf{Trion-1 combines frontier-level sequence understanding with backbone-grounded functional
reasoning.}
Trion-1 (BDLM) achieves 62.16\% accuracy on GeneralQA, closely matching
Claude Opus 4.7 (62.37\%), and 76.72\% on StructureMCQA with backbone-only
inputs. The latter remains below STELLA's reported 80.56\%~\citep{Stella},
but input modalities and test sets differ: STELLA and BioMatrix's published
structure-understanding evaluations use both sequence and structure~\citep{Stella,Biomatrix},
whereas backbone geometry alone may leave some functional questions underdetermined.

\FloatBarrier
\subsubsection{Protein Design}
\label{sec:e1.3}

\textbf{Text-to-sequence.} Table~\ref{tab:design-results} shows both the progress and remaining
limitations of our unified models in text-conditioned protein design. Without further
design-specific refinement, Trion-1 (BDLM) improves over the prior unified tri-modal baseline
BioMatrix in foldability (pLDDT: 70.65 vs.\ 60.32), functional recovery(IPRRecovery: 47.50
vs.\ 31.40), and repetition control, although its lower SR indicates less reliable production of
evaluable sequences. However, it still trails AMix2-dLLM in SR (62.18 vs.\ 69.31), pLDDT (70.65
vs.\ 74.75), and IPRRecovery (47.50 vs.\ 62.61), leaving a clear gap in functional design quality.

\suppressfloats[t]
\begin{table}[htbp]
    \centering
    \newcommand{\designOracle}{\cellcolor{gray!15}}
    \caption{\textbf{Text-conditioned protein design.}
    All metrics use a 0--100 scale. SR covers all 870 prompts; other metrics average over evaluated
    sequences.
    Yellow: our models. Gray: per-metric group best (ties included); --: unavailable; Params: total model size.
    $^\ddagger$: likelihood-ranked best-of-10; $^\dagger$: pLDDT on 819 sequences.}
    \label{tab:design-results}
    \footnotesize
    \setlength{\tabcolsep}{3.5pt}
    \renewcommand{\arraystretch}{1.08}
    \begin{tabular}{@{}l@{\hspace{3pt}}c@{\hspace{3pt}}*{6}{c}@{}}
        \toprule
        \multirow{2}{*}{\textbf{Method}} & \multirow{2}{*}{\textbf{Params}}
            & \multirow{2}{*}{\textbf{SR} $\uparrow$}
            & \multicolumn{2}{c}{\textbf{Repetition}}
            & \textbf{Foldability} & \textbf{Function} & \textbf{Novelty} \\
        \cmidrule(lr){4-5}\cmidrule(lr){6-6}\cmidrule(lr){7-7}\cmidrule(lr){8-8}
        & & & Rep2 $\downarrow$ & Rep5 $\downarrow$
            & pLDDT $\uparrow$ & IPRRecovery $\uparrow$ & Hard (Seq) $\uparrow$ \\
        \midrule
        \multicolumn{8}{l}{\textit{Frontier LLMs}} \\
        \midrule
        \quad Claude Opus 4.7 & - & 84.14 & 44.65 & \designOracle 0.65 & \designOracle 71.33 & \designOracle 65.95 & 36.54 \\
        \quad Claude Opus 4.6 & - & 90.57 & 48.44 & 1.86 & 42.17 & 25.32 & 45.91 \\
        \quad Gemini 3.1 Pro Thinking & - & \designOracle 100.00 & \designOracle 44.21 & 7.17 & 48.67 & 25.92 & 45.59 \\
        \quad GPT-5.5 & - & 99.89 & 53.19 & 3.86 & 43.95 & 25.21 & 44.21 \\
        \quad DeepSeek-V4-Pro & 1.6T & 85.06 & 50.32 & 8.14 & 45.03 & 25.14 & 51.25 \\
        \quad Qwen3.5-27B & 27B & \designOracle 100.00 & 76.95 & 63.24 & 33.47 & 0.47 & \designOracle 77.93 \\
        \quad GPT-OSS-120B & 120B & \designOracle 100.00 & 46.77 & 2.20 & 31.85 & 3.56 & 53.42 \\
        \quad GLM-5.1 & 0.8T & \designOracle 100.00 & 47.34 & 4.32 & 37.64 & 7.75 & 54.46 \\
        \midrule
        \multicolumn{8}{l}{\textit{Protein Design Models}} \\
        \midrule
        \quad CFPGen & 0.7B & 36.67 & 58.91 & 13.65 & 75.78 & 31.99 & 48.43 \\
        \quad Chroma & 0.3B & \designOracle 100.00 & 52.02 & \designOracle 0.44 & 60.46 & 0.02 & 59.69 \\
        \quad ESM3 & 1.4B & 94.94 & 69.46 & 21.64 & 60.56 & 19.20 & \designOracle 73.70 \\
        \quad Pinal$^\ddagger$ & 2B & \designOracle 100.00 & 42.47 & 3.23 & \designOracle 80.33 & \designOracle 47.78 & 38.60 \\
        \quad ProDVa$^\dagger$ & 1.8B & 99.08 & \designOracle 25.35 & 6.67 & 70.22 & 27.43 & 23.49 \\
        \quad ProteinDT & 0.7B & \designOracle 100.00 & 52.46 & 2.03 & 38.88 & 8.63 & 61.94 \\
        \midrule
        \multicolumn{8}{l}{\textit{Sequence--Text LLMs}} \\
        \midrule
        \quad AMix2-dLLM & 4B & 69.31 & 45.22 & \designOracle 1.23 & 74.75 & \designOracle 62.61 & \designOracle 30.57 \\
        \quad AMix2-AR & 4B & \designOracle 93.22 & \designOracle 44.90 & 1.61 & \designOracle 74.89 & 39.92 & 7.57 \\
        \midrule
        \multicolumn{8}{l}{\textit{Sequence--Structure--Text LLMs}} \\
        \midrule
        \quad BioMatrix & 4B & \designOracle 98.62 & 50.93 & 18.03 & 60.32 & 31.40 & 28.69 \\
        \rowcolor{yellow!15}
        \quad Trion-1 (MDLM) & 4B & 56.55 & 72.50 & 30.89 & 50.42 & 23.27 & \designOracle 72.42 \\
        \rowcolor{yellow!15}
        \quad Trion-1 (BDLM) & 4B & 62.18 & \designOracle 43.79 & \designOracle 1.63 & \designOracle 70.65 & \designOracle 47.50 & 39.29 \\
        \bottomrule
    \end{tabular}
\end{table}

Within our models, BDLM consistently improves over MDLM in foldability, functional recovery, and
repetition control. Without adaptive padding, MDLM tends to fill the inference budget, potentially
exacerbating repetitive continuations and reducing sequence quality. Such repetition is a
longstanding challenge in diffusion-based protein generation~\citep{DPLM} and persists despite the
improvements from block diffusion. 

These results indicate progress over prior unified tri-modal modeling, while highlighting the need
for further design-specific refinement in adaptive length and format control, repetition reduction,
and functional fidelity. All quality comparisons are conditional on each model's evaluated subset.

\begin{figure}[H]
    \centering
    \includegraphics[width=\linewidth]{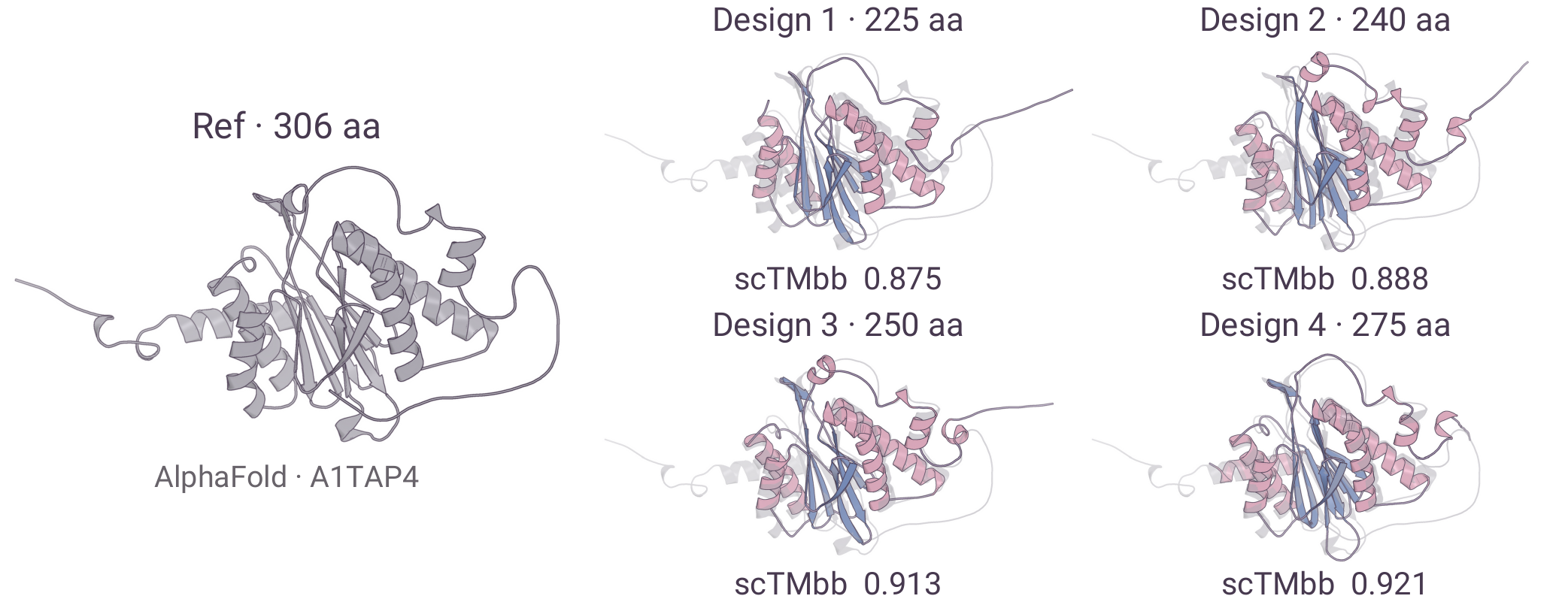}
    \caption{Text-conditioned PrcB backbone designs. The reference at left is enlarged;
    the four designs share the same scale. Pink: helices; blue: sheets; lavender: loops;
    gray: aligned AlphaFold reference. scTMbb measures ProteinMPNN--ESMFold
    self-consistency against each design backbone.}
    \label{fig:t2b-case}
\end{figure}

\Needspace{16\baselineskip}
\textbf{Text-to-backbone.} Without a suitable benchmark in our current evaluation setup, we conduct
a series of text-to-backbone case studies across multiple functional descriptions, target lengths,
and sampling seeds. Figure~\ref{fig:t2b-case} presents a representative PrcB example: the selected
designs preserve the reference-like fold with local variations and achieve scTMbb scores of
0.875--0.921, consistent with retrieval-like behavior. We hypothesize that the dataset's specific,
family-level descriptions constrain the design space; structural self-consistency alone does not
establish functional fidelity.

\FloatBarrier
\subsection{Qualitative Analysis}
\label{sec:e2}

\FloatBarrier
\subsubsection{Embedding Geometry}
\label{sec:e2.1}

\begin{figure}[!ht]
    \centering
    \begin{subfigure}[t]{0.35\linewidth}
        \centering
        \includegraphics[width=\linewidth]{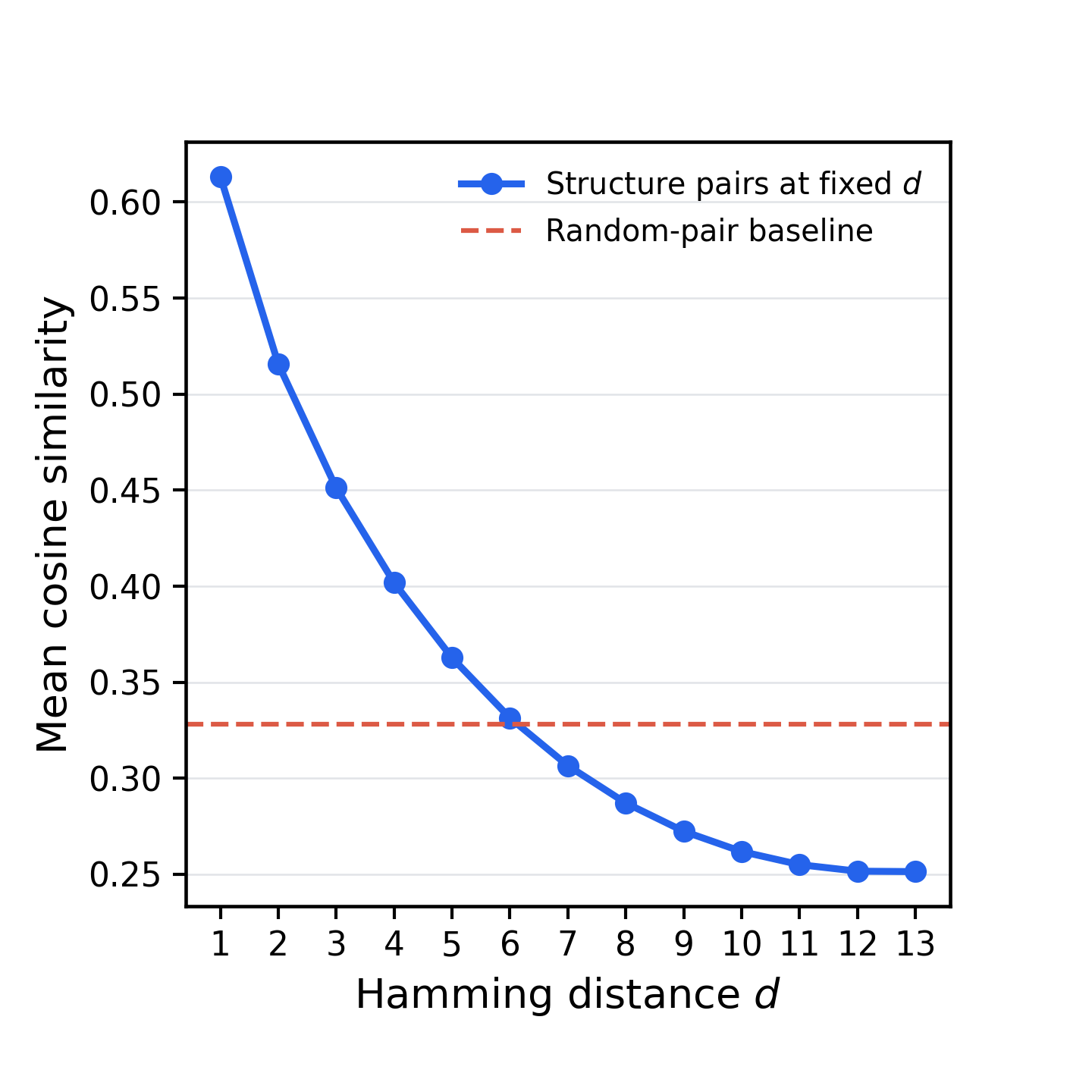}
        \caption{Similarity vs. Code distance.}
    \end{subfigure}
    \hspace{-0.01\linewidth}
    \begin{subfigure}[t]{0.35\linewidth}
        \centering
        \includegraphics[width=\linewidth]{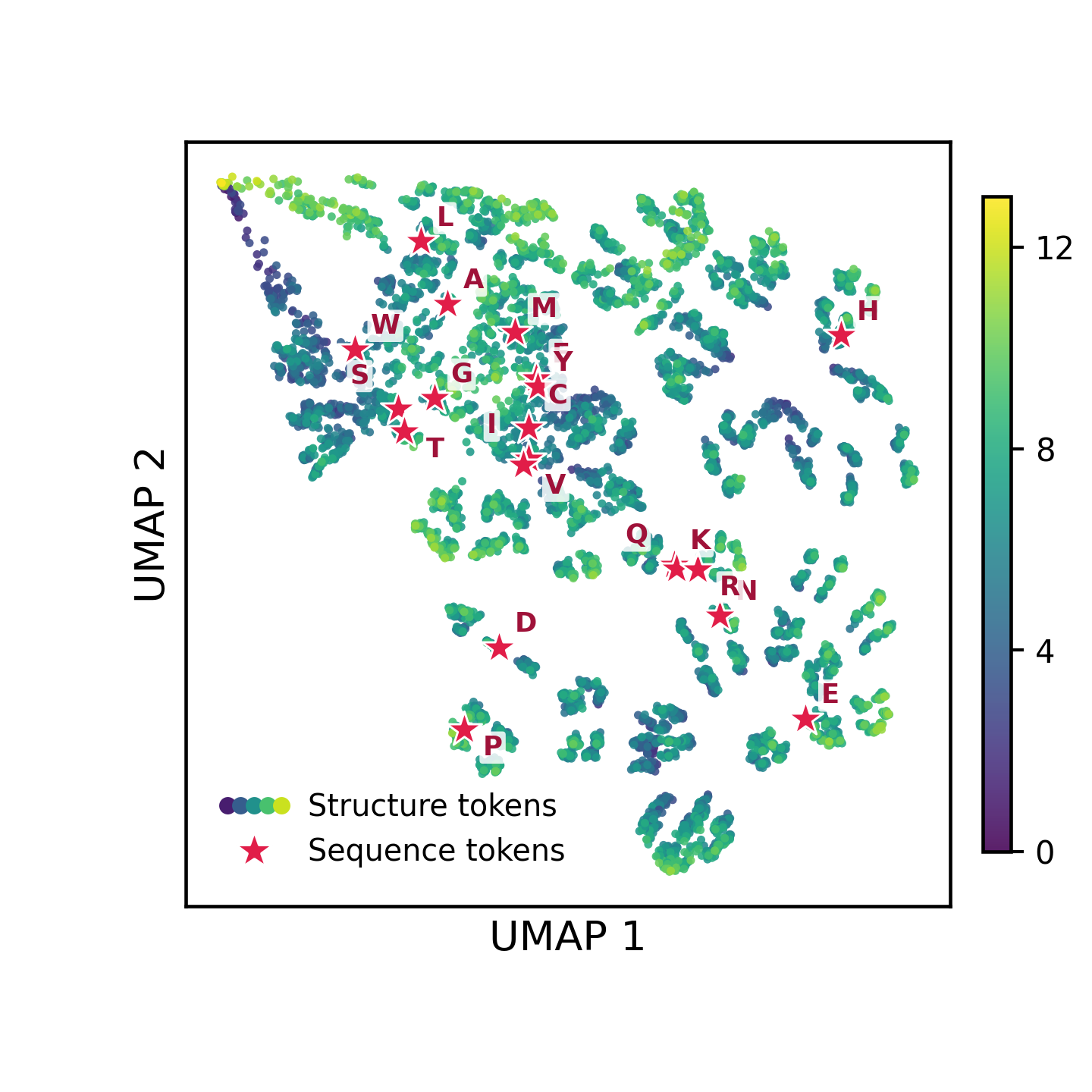}
        \caption{Visualization of embedding space.}
    \end{subfigure}

    \caption{
    \textbf{Geometric organization of the shared token space.}
    (a) Embedding similarity decreases monotonically with Hamming distance in the quantized n-bit structure-code space, as described in \ref{sec:m1}.
    (b) UMAP visualization shows that structure tokens form an organized manifold while sequence tokens occupy the same global embedding space; color intensity indicates the Hamming weight of each structure token, i.e., the number of 1s in its n-bit code.
    }
    \label{fig:emb}
\end{figure}

\textbf{Embedding geometry is preserved.} Unlike vision models that often reuse or initialize from
codebook embeddings, Trion-1 learns its structure-token embeddings from scratch. As shown in
Fig.~\ref{fig:emb}, embedding similarity decreases consistently with Hamming distance in the LFQ
code space, indicating that the tokenizer topology is largely preserved. UMAP further reveals
organized ring-like manifolds, with several amino-acid tokens near characteristic structural
neighborhoods, suggesting stable local sequence--structure correspondences. Meanwhile, structural
regions distant from any amino-acid token reflect the context-dependent, one-to-many nature of
folding.

\FloatBarrier
\subsubsection{Probing the Folding Mechanism}
\label{sec:e2.2}

\begin{figure*}[!ht]
    \centering
    \includegraphics[width=\textwidth]{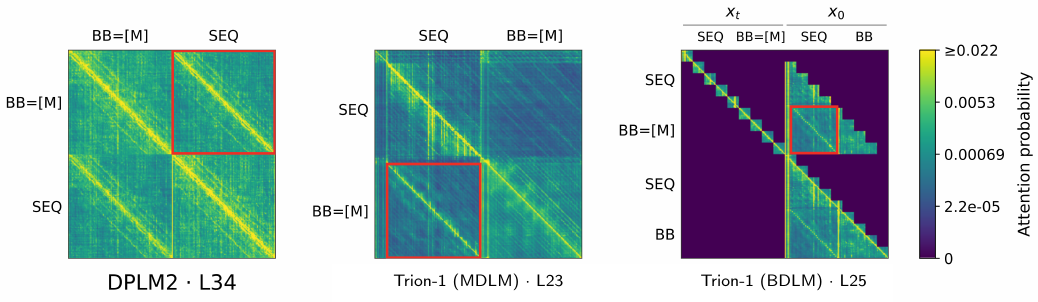}
    \caption{\textbf{Sequence--structure attention on 6WEB.} Head-averaged maps for DPLM2,
    Trion-1 (MDLM), and Trion-1 (BDLM). Red boxes mark structure queries attending to
    sequence keys. Each layer is selected by maximal conditional attention mass within $\pm 2$
    residues of the corresponding sequence position.}
    \label{fig:folding-attention}
\end{figure*}

\textbf{Folding involves residue-level translation.}
We initially attributed DPLM2's diagonal sequence--structure attention to its aligned positional
encoding. Yet similar patterns persist under global positional encoding
(Figure~\ref{fig:folding-attention}), suggesting that residue-level translation can emerge from the
folding task itself. This motivates a conceptual pathway: $\mathrm{sequence} \rightarrow
\mathrm{contact\ constraints} \rightarrow \mathrm{structure}$.

\textbf{Causal generation requires harder anticipation.}
For contacting residues, causal masking prevents the earlier structure token from directly accessing
its partner's future structural state. Early next-token decisions must therefore anticipate nonlocal
constraints. This is consistent with weak BioMatrix@1 but better BioMatrix@200.

\FloatBarrier
\section{Limitation and Future Direction}
Trion-1 currently focuses on protein-only modeling and does not yet cover richer biomolecular
systems such as complexes, ligands, or nucleic acids. Moreover, despite strong performance across
diverse tasks, a gap remains to the strongest task-specialized models, particularly on
structure-intensive problems such as protein folding.

Looking forward, we see two promising directions. (1) Further improving Trion-1's capability
ceiling through multi-codebook structure representations, stronger structure decoders \citep{DPLM2,
DPLM2.1}, and targeted augmentation for short proteins and challenging cases. (2) Exploiting
interleaved generation for test-time scaling, where composable capabilities are iteratively
generated, evaluated, and refined with additional inference-time computation. With sufficiently
strong base capabilities, UniProt-scale sequence--structure--function corpora could be available,
providing the substrate for closed-loop self-improvement.

\FloatBarrier
\section{Conclusion}
We presented Trion-1, a native unified tri-modal diffusion language model that brings protein
sequence, structure, and natural language into a single generative framework. By casting six core
capabilities across the sequence--structure--function space as conditional generation, Trion-1
achieves broadly competitive performance at the 4B scale while substantially strengthening the
folding capability of unified protein models, demonstrating that strong structure modeling can
coexist with broad understanding, generation, and native language interaction within a single model.

Our analyses further support the importance of non-causal modeling for protein structure and show
that the geometric organization induced by the structure tokenizer can be naturally absorbed into
the language-model embedding space. More broadly, Trion-1 turns specialized protein capabilities
into composable generative primitives, enabling interleaved workflows and providing a foundation for
future test-time scaling and closed-loop self-improvement through self-generated data.

\clearpage
\section*{Contributions}

\textbf{Project Lead}

Tianyu Lu\textsuperscript{1,2,3}

\medskip
\textbf{Contributors}

Kaicheng Yang\textsuperscript{4},
Jiangtao Feng\textsuperscript{1,2},
Keyue Qiu\textsuperscript{1,2,3},
Yawen Ouyang\textsuperscript{1,3},
Yixin Wu\textsuperscript{1,3,5},
Xinbo Zhang\textsuperscript{1,3},
Lihao Wang\textsuperscript{1,3},
Wei-Ying Ma\textsuperscript{2,6}

\medskip
\textbf{Correspondence}

Hao Zhou\textsuperscript{1,2,3}\\
\href{mailto:zhouhao@air.tsinghua.edu.cn}{\nolinkurl{zhouhao@air.tsinghua.edu.cn}}

\section*{Affiliations}

\textsuperscript{1}Generative Symbolic Intelligence Lab (GenSI), Tsinghua University

\textsuperscript{2}Institute for AI Industry Research (AIR), Tsinghua University

\textsuperscript{3}Shanghai Artificial Intelligence Laboratory

\textsuperscript{4}Shanghai Jiao Tong University

\textsuperscript{5}Fudan University

\textsuperscript{6}City University of Hong Kong

\section*{Acknowledgments}
We thank Zaixiang Zheng and Xinyou Wang for their guidance and valuable suggestions
throughout this project, and Yifei Wang for assistance with data synthesis.
This work is supported by Shanghai Artificial Intelligence Laboratory.

\clearpage

\bibliography{references}

\begin{thebibliography}{65}
\providecommand{\natexlab}[1]{#1}
\providecommand{\url}[1]{\texttt{#1}}
\expandafter\ifx\csname urlstyle\endcsname\relax
  \providecommand{\doi}[1]{doi: #1}\else
  \providecommand{\doi}{doi: \begingroup \urlstyle{rm}\Url}\fi

\bibitem[Rives et~al.(2021)Rives, Meier, Sercu, Goyal, Lin, Liu, Guo, Ott, Zitnick, Ma, and Fergus]{ESM2}
Alexander Rives, Joshua Meier, Tom Sercu, Siddharth Goyal, Zeming Lin, Jason Liu, Demi Guo, Myle Ott, C.~Lawrence Zitnick, Jerry Ma, and Rob Fergus.
\newblock Biological structure and function emerge from scaling unsupervised learning to 250 million protein sequences.
\newblock \emph{Proceedings of the National Academy of Sciences}, 118\penalty0 (15):\penalty0 e2016239118, 2021.
\newblock \doi{10.1073/pnas.2016239118}.
\newblock URL \url{https://www.pnas.org/doi/abs/10.1073/pnas.2016239118}.

\bibitem[Dill and MacCallum(2012)]{Structure}
Ken~A. Dill and Justin~L. MacCallum.
\newblock The protein-folding problem, 50 years on.
\newblock \emph{Science}, 338\penalty0 (6110):\penalty0 1042--1046, 2012.
\newblock \doi{10.1126/science.1219021}.
\newblock URL \url{https://www.science.org/doi/abs/10.1126/science.1219021}.

\bibitem[gen(2021)]{gene2021}
The gene ontology resource: enriching a gold mine.
\newblock \emph{Nucleic acids research}, 49\penalty0 (D1):\penalty0 D325--D334, 2021.

\bibitem[uni(2023)]{uniprot2023}
Uniprot: the universal protein knowledgebase in 2023.
\newblock \emph{Nucleic acids research}, 51\penalty0 (D1):\penalty0 D523--D531, 2023.

\bibitem[Ma et~al.(2025)Ma, Fan, Wang, Chen, Lin, Li, Feng, Cao, Zhang, and Gao]{ProtTex}
Zicheng Ma, Chuanliu Fan, Zhicong Wang, Zhenyu Chen, Xiaohan Lin, Yanheng Li, Shihao Feng, Ziqiang Cao, Jun Zhang, and Yi~Qin Gao.
\newblock Prottex: Structure-in-context reasoning and editing of proteins with large language models.
\newblock \emph{Journal of Chemical Information and Modeling}, 65\penalty0 (13):\penalty0 6599--6612, 2025.

\bibitem[Chen et~al.(2024)Chen, Chen, Xie, Xue, Zhang, Zhou, and Fang]{HelixProtX}
Zhiyuan Chen, Tianhao Chen, Chenggang Xie, Yang Xue, Xiaonan Zhang, Jingbo Zhou, and Xiaomin Fang.
\newblock Unifying sequences, structures, and descriptions for any-to-any protein generation with the large multimodal model helixprotx.
\newblock \emph{arXiv preprint arXiv:2407.09274}, 2024.

\bibitem[Team(2025)]{Chameleon}
Chameleon Team.
\newblock Chameleon: Mixed-modal early-fusion foundation models, 2025.
\newblock URL \url{https://arxiv.org/abs/2405.09818}.

\bibitem[Wang et~al.(2024{\natexlab{a}})Wang, Zhang, Luo, Sun, Cui, Wang, Zhang, Wang, Li, Yu, Zhao, Ao, Min, Li, Wu, Zhao, Zhang, Wang, Liu, He, Yang, Liu, Lin, Huang, and Wang]{Emu3}
Xinlong Wang, Xiaosong Zhang, Zhengxiong Luo, Quan Sun, Yufeng Cui, Jinsheng Wang, Fan Zhang, Yueze Wang, Zhen Li, Qiying Yu, Yingli Zhao, Yulong Ao, Xuebin Min, Tao Li, Boya Wu, Bo~Zhao, Bowen Zhang, Liangdong Wang, Guang Liu, Zheqi He, Xi~Yang, Jingjing Liu, Yonghua Lin, Tiejun Huang, and Zhongyuan Wang.
\newblock Emu3: Next-token prediction is all you need, 2024{\natexlab{a}}.
\newblock URL \url{https://arxiv.org/abs/2409.18869}.

\bibitem[{Anthropic}(2026)]{Claude47}
{Anthropic}.
\newblock Claude opus 4.7 system card.
\newblock Technical report, Anthropic, 2026.

\bibitem[Qiu et~al.(2026)Qiu, Wu, Wang, Ouyang, Yu, Zhou, Lv, Xue, Song, Zhang, et~al.]{Amix2}
Keyue Qiu, Yixin Wu, Lihao Wang, Yawen Ouyang, Jixiang Yu, Zihan Zhou, Changze Lv, Dongyu Xue, Yuxuan Song, Xinbo Zhang, et~al.
\newblock Amix-2: Establishing protein as a native modality in large language models.
\newblock \emph{arXiv preprint arXiv:2605.30963}, 2026.

\bibitem[Bie et~al.(2026)Bie, Cai, Cao, Cao, Chen, Chen, Chen, Ding, Du, Gong, et~al.]{Llada2.2}
Tiwei Bie, Jiansheng Cai, Maosong Cao, Xiang Cao, Binbin Chen, Fuyuan Chen, Kun Chen, Yuqi Ding, Lun Du, Zhuocheng Gong, et~al.
\newblock Llada2. 2: Enabling agentic diffusion language models via levenshtein editing.
\newblock Technical report, CSPaper OpenPrint, 2026.

\bibitem[Team et~al.(2026)Team, Ta{\"\i}ga, Assiene, Calandriello, Chaabouni, Gante, von Glehn, Keating, Knutsen, Kukla, et~al.]{DiffusionGemma}
DiffusionGemma Team, Adrien~Ali Ta{\"\i}ga, James Assiene, Daniele Calandriello, Rahma Chaabouni, Jo{\~a}o Gante, Tamara von Glehn, Nate Keating, Chris Knutsen, Martin Kukla, et~al.
\newblock Diffusiongemma technical report.
\newblock \emph{arXiv preprint arXiv:2608.00146}, 2026.

\bibitem[Sahoo et~al.(2024)Sahoo, Arriola, Schiff, Gokaslan, Marroquin, Chiu, Rush, and Kuleshov]{MDLM}
Subham~Sekhar Sahoo, Marianne Arriola, Yair Schiff, Aaron Gokaslan, Edgar Marroquin, Justin~T. Chiu, Alexander Rush, and Volodymyr Kuleshov.
\newblock Simple and {Effective} {Masked} {Diffusion} {Language} {Models}, November 2024.
\newblock URL \url{http://arxiv.org/abs/2406.07524}.
\newblock arXiv:2406.07524 [cs].

\bibitem[Shi et~al.(2025)Shi, Han, Wang, Doucet, and Titsias]{MD4}
Jiaxin Shi, Kehang Han, Zhe Wang, Arnaud Doucet, and Michalis~K. Titsias.
\newblock Simplified and generalized masked diffusion for discrete data, 2025.
\newblock URL \url{https://arxiv.org/abs/2406.04329}.

\bibitem[Ou et~al.(2026)Ou, Nie, Xue, Zhu, Sun, Li, and Li]{RADD}
Jingyang Ou, Shen Nie, Kaiwen Xue, Fengqi Zhu, Jiacheng Sun, Zhenguo Li, and Chongxuan Li.
\newblock Your absorbing discrete diffusion secretly models the conditional distributions of clean data, 2026.
\newblock URL \url{https://arxiv.org/abs/2406.03736}.

\bibitem[Arriola et~al.(2025)Arriola, Gokaslan, Chiu, Yang, Qi, Han, Sahoo, and Kuleshov]{BD3LM}
Marianne Arriola, Aaron Gokaslan, Justin~T. Chiu, Zhihan Yang, Zhixuan Qi, Jiaqi Han, Subham~Sekhar Sahoo, and Volodymyr Kuleshov.
\newblock Block diffusion: Interpolating between autoregressive and diffusion language models, 2025.
\newblock URL \url{https://arxiv.org/abs/2503.09573}.

\bibitem[Zhou et~al.(2026)Zhou, Chen, Tong, and Song]{dllm}
Zhanhui Zhou, Lingjie Chen, Hanghang Tong, and Dawn Song.
\newblock {dLLM}: {Simple} {Diffusion} {Language} {Modeling}, February 2026.
\newblock URL \url{http://arxiv.org/abs/2602.22661}.
\newblock arXiv:2602.22661 [cs] version: 1.

\bibitem[Bie et~al.(2025)Bie, Cao, Chen, Du, Gong, Gong, Gu, Hu, Huang, Lan, et~al.]{Llada2.0}
Tiwei Bie, Maosong Cao, Kun Chen, Lun Du, Mingliang Gong, Zhuochen Gong, Yanmei Gu, Jiaqi Hu, Zenan Huang, Zhenzhong Lan, et~al.
\newblock Llada2. 0: Scaling up diffusion language models to 100b.
\newblock \emph{arXiv preprint arXiv:2512.15745}, 2025.

\bibitem[Zhu et~al.(2026)Zhu, Shi, Ge, Tan, Xu, Zhu, Kuen, Goswami, Jain, Chen, Tao, and Gu]{Flare}
Yuchen Zhu, Jing Shi, Chongjian Ge, Hao Tan, Yiran Xu, Wanrong Zhu, Jason Kuen, Koustava Goswami, Rajiv Jain, Yongxin Chen, Molei Tao, and Jiuxiang Gu.
\newblock {FLARE}: {Diffusion} for {Hybrid} {Language} {Model}, 2026.
\newblock URL \url{https://arxiv.org/abs/2606.01774}.
\newblock Version Number: 1.

\bibitem[Yang et~al.(2023)Yang, Zeng, Zhao, and Chen]{AF2}
Zhenyu Yang, Xiaoxi Zeng, Yi~Zhao, and Runsheng Chen.
\newblock Alphafold2 and its applications in the fields of biology and medicine.
\newblock \emph{Signal transduction and targeted therapy}, 8\penalty0 (1):\penalty0 115, 2023.

\bibitem[Abramson et~al.(2024)Abramson, Adler, Dunger, Evans, Green, Pritzel, Ronneberger, Willmore, Ballard, Bambrick, et~al.]{AF3}
Josh Abramson, Jonas Adler, Jack Dunger, Richard Evans, Tim Green, Alexander Pritzel, Olaf Ronneberger, Lindsay Willmore, Andrew~J Ballard, Joshua Bambrick, et~al.
\newblock Accurate structure prediction of biomolecular interactions with alphafold 3.
\newblock \emph{Nature}, 630\penalty0 (8016):\penalty0 493--500, 2024.

\bibitem[Zhou et~al.(2025)Zhou, Lu, Ma, Qu, Ye, Zhang, Wang, Gui, and Gu]{SeedFold}
Yi~Zhou, Chan Lu, Yiming Ma, Wei Qu, Fei Ye, Kexin Zhang, Lan Wang, Minrui Gui, and Quanquan Gu.
\newblock Seedfold: Scaling biomolecular structure prediction.
\newblock \emph{arXiv preprint arXiv:2512.24354}, 2025.

\bibitem[Gong et~al.(2025)Gong, Chen, Zhang, Song, Zhou, and Xiao]{Protenix-mini}
Chengyue Gong, Xinshi Chen, Yuxuan Zhang, Yuxuan Song, Hao Zhou, and Wenzhi Xiao.
\newblock Protenix-mini: Efficient structure predictor via compact architecture, few-step diffusion and switchable plm.
\newblock \emph{arXiv preprint arXiv:2507.11839}, 2025.

\bibitem[Qiang et~al.(2025)Qiang, Gong, Chen, Zhang, and Xiao]{Protenix-mini+}
Bo~Qiang, Chengyue Gong, Xinshi Chen, Yuxuan Zhang, and Wenzhi Xiao.
\newblock Protenix-mini+: efficient structure prediction model with scalable pairformer.
\newblock \emph{arXiv preprint arXiv:2510.12842}, 2025.

\bibitem[Wohlwend et~al.(2025)Wohlwend, Corso, Passaro, Getz, Reveiz, Leidal, Swiderski, Atkinson, Portnoi, Chinn, et~al.]{Boltz1}
Jeremy Wohlwend, Gabriele Corso, Saro Passaro, Noah Getz, Mateo Reveiz, Ken Leidal, Wojtek Swiderski, Liam Atkinson, Tally Portnoi, Itamar Chinn, et~al.
\newblock Boltz-1 democratizing biomolecular interaction modeling.
\newblock \emph{BioRxiv}, pages 2024--11, 2025.

\bibitem[Passaro et~al.(2025)Passaro, Corso, Wohlwend, Reveiz, Thaler, Somnath, Getz, Portnoi, Roy, Stark, et~al.]{Boltz2}
Saro Passaro, Gabriele Corso, Jeremy Wohlwend, Mateo Reveiz, Stephan Thaler, Vignesh~Ram Somnath, Noah Getz, Tally Portnoi, Julien Roy, Hannes Stark, et~al.
\newblock Boltz-2: Towards accurate and efficient binding affinity prediction.
\newblock \emph{BioRxiv}, 2025.

\bibitem[team et~al.(2024)team, Boitreaud, Dent, McPartlon, Meier, Reis, Rogozhonikov, and Wu]{Chai1}
Chai~Discovery team, Jacques Boitreaud, Jack Dent, Matthew McPartlon, Joshua Meier, Vinicius Reis, Alex Rogozhonikov, and Kevin Wu.
\newblock Chai-1: Decoding the molecular interactions of life.
\newblock \emph{BioRxiv}, pages 2024--10, 2024.

\bibitem[Team et~al.(2025)Team, Boitreaud, Dent, Geisz, McPartlon, Meier, Qiao, Rogozhnikov, Rollins, Wollenhaupt, et~al.]{Chai2}
Chai~Discovery Team, Jacques Boitreaud, Jack Dent, Danny Geisz, Matthew McPartlon, Joshua Meier, Zhuoran Qiao, Alex Rogozhnikov, Nathan Rollins, Paul Wollenhaupt, et~al.
\newblock Zero-shot antibody design in a 24-well plate.
\newblock \emph{BioRxiv}, pages 2025--07, 2025.

\bibitem[Wang et~al.(2026)Wang, Lu, Jaitly, Susskind, and Bautista]{SimpleFold}
Yuyang Wang, Jiarui Lu, Navdeep Jaitly, Joshua Susskind, and Miguel~Angel Bautista.
\newblock Simplefold: Folding proteins is simpler than you think.
\newblock In \emph{International Conference on Learning Representations}, volume 2026, pages 32027--32054, 2026.

\bibitem[Dauparas et~al.(2022)Dauparas, Anishchenko, Bennett, Bai, Ragotte, Milles, Wicky, Courbet, de~Haas, Bethel, et~al.]{PMPNN}
Justas Dauparas, Ivan Anishchenko, Nathaniel Bennett, Hua Bai, Robert~J Ragotte, Lukas~F Milles, Basile~IM Wicky, Alexis Courbet, Rob~J de~Haas, Neville Bethel, et~al.
\newblock Robust deep learning--based protein sequence design using proteinmpnn.
\newblock \emph{Science}, 378\penalty0 (6615):\penalty0 49--56, 2022.

\bibitem[Dauparas et~al.(2025)Dauparas, Lee, Pecoraro, An, Anishchenko, Glasscock, and Baker]{LigandMPNN}
Justas Dauparas, Gyu~Rie Lee, Robert Pecoraro, Linna An, Ivan Anishchenko, Cameron Glasscock, and David Baker.
\newblock Atomic context-conditioned protein sequence design using ligandmpnn.
\newblock \emph{Nature Methods}, 22\penalty0 (4):\penalty0 717--723, 2025.

\bibitem[Liu et~al.(2025{\natexlab{a}})Liu, Wu, Wang, Wang, Chen, Li, Chen, and Liu]{ABACUS}
Yufeng Liu, Rui Wu, Xinyu Wang, Sheng Wang, Linghui Chen, Fudong Li, Quan Chen, and Haiyan Liu.
\newblock Enhancing functional proteins through multimodal inverse folding with abacus-t.
\newblock \emph{Nature Communications}, 16\penalty0 (1):\penalty0 10177, 2025{\natexlab{a}}.

\bibitem[Elnaggar et~al.(2021)Elnaggar, Heinzinger, Dallago, Rehawi, Wang, Jones, Gibbs, Feher, Angerer, Steinegger, et~al.]{ProtT5}
Ahmed Elnaggar, Michael Heinzinger, Christian Dallago, Ghalia Rehawi, Yu~Wang, Llion Jones, Tom Gibbs, Tamas Feher, Christoph Angerer, Martin Steinegger, et~al.
\newblock Prottrans: toward understanding the language of life through self-supervised learning.
\newblock \emph{IEEE transactions on pattern analysis and machine intelligence}, 44\penalty0 (10):\penalty0 7112--7127, 2021.

\bibitem[Xu et~al.(2023)Xu, Yuan, Miret, and Tang]{ProtST}
Minghao Xu, Xinyu Yuan, Santiago Miret, and Jian Tang.
\newblock Protst: Multi-modality learning of protein sequences and biomedical texts.
\newblock In \emph{International conference on machine learning}, pages 38749--38767. PMLR, 2023.

\bibitem[Jing et~al.(2020)Jing, Eismann, Suriana, Townshend, and Dror]{GVP}
Bowen Jing, Stephan Eismann, Patricia Suriana, Raphael~JL Townshend, and Ron Dror.
\newblock Learning from protein structure with geometric vector perceptrons.
\newblock \emph{arXiv preprint arXiv:2009.01411}, 2020.

\bibitem[Zhang et~al.(2022)Zhang, Xu, Jamasb, Chenthamarakshan, Lozano, Das, and Tang]{GearNet}
Zuobai Zhang, Minghao Xu, Arian Jamasb, Vijil Chenthamarakshan, Aurelie Lozano, Payel Das, and Jian Tang.
\newblock Protein representation learning by geometric structure pretraining.
\newblock \emph{arXiv preprint arXiv:2203.06125}, 2022.

\bibitem[Madani et~al.(2020)Madani, McCann, Naik, Keskar, Anand, Eguchi, Huang, and Socher]{Progen}
Ali Madani, Bryan McCann, Nikhil Naik, Nitish~Shirish Keskar, Namrata Anand, Raphael~R Eguchi, Po-Ssu Huang, and Richard Socher.
\newblock Progen: Language modeling for protein generation.
\newblock \emph{BioRxiv}, pages 2020--03, 2020.

\bibitem[Liu et~al.(2025{\natexlab{b}})Liu, Li, Li, Gitter, Zhu, Lu, Xu, Nie, Ramanathan, Xiao, et~al.]{ProteinDT}
Shengchao Liu, Yanjing Li, Zhuoxinran Li, Anthony Gitter, Yutao Zhu, Jiarui Lu, Zhao Xu, Weili Nie, Arvind Ramanathan, Chaowei Xiao, et~al.
\newblock A text-guided protein design framework.
\newblock \emph{Nature Machine Intelligence}, 7\penalty0 (4):\penalty0 580--591, 2025{\natexlab{b}}.

\bibitem[Guo et~al.(2025)Guo, Li, Liu, Pan, and Shen]{ProtDAT}
Xiao-Yu Guo, Yi-Fan Li, Yuan Liu, Xiaoyong Pan, and Hong-Bin Shen.
\newblock Ab-initio amino acid sequence design from protein text description with protdat.
\newblock \emph{Nature Communications}, 16\penalty0 (1):\penalty0 10544, 2025.

\bibitem[Alamdari et~al.(2023)Alamdari, Thakkar, Van Den~Berg, Tenenholtz, Strome, Moses, Lu, Fusi, Amini, and Yang]{EvoDiff}
Sarah Alamdari, Nitya Thakkar, Rianne Van Den~Berg, Neil Tenenholtz, Robert Strome, Alan~M Moses, Alex~X Lu, Nicol{\`o} Fusi, Ava~P Amini, and Kevin~K Yang.
\newblock Protein generation with evolutionary diffusion: sequence is all you need.
\newblock \emph{BioRxiv}, pages 2023--09, 2023.

\bibitem[Wang et~al.(2024{\natexlab{b}})Wang, Zheng, Ye, Xue, Huang, and Gu]{DPLM}
Xinyou Wang, Zaixiang Zheng, Fei Ye, Dongyu Xue, Shujian Huang, and Quanquan Gu.
\newblock Diffusion language models are versatile protein learners.
\newblock \emph{arXiv preprint arXiv:2402.18567}, 2024{\natexlab{b}}.

\bibitem[Watson et~al.(2023)Watson, Juergens, Bennett, Trippe, Yim, Eisenach, Ahern, Borst, Ragotte, Milles, et~al.]{RFDiffusion}
Joseph~L Watson, David Juergens, Nathaniel~R Bennett, Brian~L Trippe, Jason Yim, Helen~E Eisenach, Woody Ahern, Andrew~J Borst, Robert~J Ragotte, Lukas~F Milles, et~al.
\newblock De novo design of protein structure and function with rfdiffusion.
\newblock \emph{Nature}, 620\penalty0 (7976):\penalty0 1089--1100, 2023.

\bibitem[Ahern et~al.(2026)Ahern, Yim, Tischer, Salike, Woodbury, Kim, Kalvet, Kipnis, Coventry, Altae-Tran, et~al.]{RFDiffusion2}
Woody Ahern, Jason Yim, Doug Tischer, Saman Salike, Seth~M Woodbury, Donghyo Kim, Indrek Kalvet, Yakov Kipnis, Brian Coventry, Han~Raut Altae-Tran, et~al.
\newblock Atom-level enzyme active site scaffolding using rfdiffusion2.
\newblock \emph{Nature Methods}, 23\penalty0 (1):\penalty0 96--105, 2026.

\bibitem[Geffner et~al.(2025)Geffner, Didi, Zhang, Reidenbach, Cao, Yim, Geiger, Dallago, Kucukbenli, Vahdat, et~al.]{Proteina}
Tomas Geffner, Kieran Didi, Zuobai Zhang, Danny Reidenbach, Zhonglin Cao, Jason Yim, Mario Geiger, Christian Dallago, Emine Kucukbenli, Arash Vahdat, et~al.
\newblock Proteina: Scaling flow-based protein structure generative models.
\newblock In \emph{International Conference on Learning Representations}, volume 2025, pages 98803--98851, 2025.

\bibitem[Geffner et~al.(2026)Geffner, Didi, Cao, Reidenbach, Zhang, Dallago, Kucukbenli, Kreis, and Vahdat]{LaProteina}
Tomas Geffner, Kieran Didi, Zhonglin Cao, Danny Reidenbach, Zuobai Zhang, Christian Dallago, Emine Kucukbenli, Karsten Kreis, and Arash Vahdat.
\newblock La-proteina: Atomistic protein generation via partially latent flow matching.
\newblock In \emph{International Conference on Learning Representations}, volume 2026, pages 67350--67395, 2026.

\bibitem[Su et~al.(2025)Su, He, You, Jiang, Zhou, Zhang, Wang, Su, Tolstoy, Chang, et~al.]{ProTrek}
Jin Su, Yan He, Shiyang You, Shiyu Jiang, Xibin Zhou, Xuting Zhang, Yuxuan Wang, Xining Su, Igor Tolstoy, Xing Chang, et~al.
\newblock A trimodal protein language model enables advanced protein searches.
\newblock \emph{Nature Biotechnology}, pages 1--7, 2025.

\bibitem[Zhang et~al.(2024)Zhang, Guo, Schaffer, Ko, Singh, Rahmani, Grotjahn, Villa, Gilson, Wang, et~al.]{ProteinAligner}
Li~Zhang, Han Guo, Leah Schaffer, Young~Su Ko, Digvijay Singh, Hamid Rahmani, Danielle Grotjahn, Elizabeth Villa, Michael Gilson, Wei Wang, et~al.
\newblock Proteinaligner: A multi-modal pretraining framework for protein foundation models.
\newblock \emph{bioRxiv}, pages 2024--10, 2024.

\bibitem[Hayes et~al.(2025)Hayes, Rao, Akin, Sofroniew, Oktay, Lin, Verkuil, Tran, Deaton, Wiggert, et~al.]{ESM3}
Thomas Hayes, Roshan Rao, Halil Akin, Nicholas~J Sofroniew, Deniz Oktay, Zeming Lin, Robert Verkuil, Vincent~Q Tran, Jonathan Deaton, Marius Wiggert, et~al.
\newblock Simulating 500 million years of evolution with a language model.
\newblock \emph{Science}, 387\penalty0 (6736):\penalty0 850--858, 2025.

\bibitem[Pei et~al.(2026)Pei, Zhou, Duan, Zhao, Li, Guo, He, Li, Hsieh, He, et~al.]{Biomatrix}
Qizhi Pei, Zhimeng Zhou, Yi~Duan, Yiyang Zhao, Wei Li, Han Guo, Liang He, Chengping Li, Chang-Yu Hsieh, Conghui He, et~al.
\newblock Biomatrix: Towards a comprehensive biological foundation model spanning the modality matrix of sequences, structures, and language.
\newblock \emph{arXiv preprint arXiv:2606.22138}, 2026.

\bibitem[Yang et~al.(2025)Yang, Li, Yang, Zhang, Hui, Zheng, Yu, Gao, Huang, Lv, Zheng, Liu, Zhou, Huang, Hu, Ge, Wei, Lin, Tang, Yang, Tu, Zhang, Yang, Yang, Zhou, Zhou, Lin, Dang, Bao, Yang, Yu, Deng, Li, Xue, Li, Zhang, Wang, Zhu, Men, Gao, Liu, Luo, Li, Tang, Yin, Ren, Wang, Zhang, Ren, Fan, Su, Zhang, Zhang, Wan, Liu, Wang, Cui, Zhang, Zhou, and Qiu]{Qwen3}
An~Yang, Anfeng Li, Baosong Yang, Beichen Zhang, Binyuan Hui, Bo~Zheng, Bowen Yu, Chang Gao, Chengen Huang, Chenxu Lv, Chujie Zheng, Dayiheng Liu, Fan Zhou, Fei Huang, Feng Hu, Hao Ge, Haoran Wei, Huan Lin, Jialong Tang, Jian Yang, Jianhong Tu, Jianwei Zhang, Jianxin Yang, Jiaxi Yang, Jing Zhou, Jingren Zhou, Junyang Lin, Kai Dang, Keqin Bao, Kexin Yang, Le~Yu, Lianghao Deng, Mei Li, Mingfeng Xue, Mingze Li, Pei Zhang, Peng Wang, Qin Zhu, Rui Men, Ruize Gao, Shixuan Liu, Shuang Luo, Tianhao Li, Tianyi Tang, Wenbiao Yin, Xingzhang Ren, Xinyu Wang, Xinyu Zhang, Xuancheng Ren, Yang Fan, Yang Su, Yichang Zhang, Yinger Zhang, Yu~Wan, Yuqiong Liu, Zekun Wang, Zeyu Cui, Zhenru Zhang, Zhipeng Zhou, and Zihan Qiu.
\newblock Qwen3 technical report, 2025.
\newblock URL \url{https://arxiv.org/abs/2505.09388}.

\bibitem[Van~Kempen et~al.(2024)Van~Kempen, Kim, Tumescheit, Mirdita, Lee, Gilchrist, S{\"o}ding, and Steinegger]{FoldSeek}
Michel Van~Kempen, Stephanie~S Kim, Charlotte Tumescheit, Milot Mirdita, Jeongjae Lee, Cameron~LM Gilchrist, Johannes S{\"o}ding, and Martin Steinegger.
\newblock Fast and accurate protein structure search with foldseek.
\newblock \emph{Nature biotechnology}, 42\penalty0 (2):\penalty0 243--246, 2024.

\bibitem[Kingma and Welling(2022)]{VAE}
Diederik~P Kingma and Max Welling.
\newblock Auto-encoding variational bayes, 2022.
\newblock URL \url{https://arxiv.org/abs/1312.6114}.

\bibitem[van~den Oord et~al.(2018)van~den Oord, Vinyals, and Kavukcuoglu]{VQVAE}
Aaron van~den Oord, Oriol Vinyals, and Koray Kavukcuoglu.
\newblock Neural discrete representation learning, 2018.
\newblock URL \url{https://arxiv.org/abs/1711.00937}.

\bibitem[Shaulov et~al.(2026)Shaulov, Hazan, Wolf, and Chefer]{FlowMo}
Ariel Shaulov, Itay Hazan, Lior Wolf, and Hila Chefer.
\newblock Flowmo: Variance-based flow guidance for coherent motion in video generation.
\newblock \emph{Advances in Neural Information Processing Systems}, 38:\penalty0 82725--82754, 2026.

\bibitem[Chen et~al.(2025)Chen, Girdhar, Wang, Rambhatla, and Misra]{DiTo}
Yinbo Chen, Rohit Girdhar, Xiaolong Wang, Sai~Saketh Rambhatla, and Ishan Misra.
\newblock Diffusion autoencoders are scalable image tokenizers.
\newblock \emph{arXiv preprint arXiv:2501.18593}, 2025.

\bibitem[Dilip et~al.(2026{\natexlab{a}})Dilip, Zhang, Varshney, and Van~Valen]{Kanzi}
Rohit Dilip, Evan Zhang, Ayush Varshney, and David Van~Valen.
\newblock Flow autoencoders are effective protein tokenizers.
\newblock In \emph{International Conference on Learning Representations}, volume 2026, pages 145901--145926, 2026{\natexlab{a}}.

\bibitem[Giri et~al.(2026)Giri, Farrell, and Bouchard]{Yeti}
Nabin Giri, Steven Farrell, and Kristofer~E Bouchard.
\newblock Yeti: A compact protein structure tokenizer for reconstruction and multi-modal generation.
\newblock \emph{arXiv preprint arXiv:2605.09981}, 2026.

\bibitem[Dilip et~al.(2026{\natexlab{b}})Dilip, Varshney, and Valen]{Adaptive}
Rohit Dilip, Ayush Varshney, and David~Van Valen.
\newblock Adaptive {Protein} {Tokenization}, February 2026{\natexlab{b}}.
\newblock URL \url{http://arxiv.org/abs/2602.06418}.
\newblock arXiv:2602.06418 [cs].

\bibitem[Yu et~al.(2024)Yu, Lezama, Gundavarapu, Versari, Sohn, Minnen, Cheng, Birodkar, Gupta, Gu, Hauptmann, Gong, Yang, Essa, Ross, and Jiang]{LFQ}
Lijun Yu, José Lezama, Nitesh~B. Gundavarapu, Luca Versari, Kihyuk Sohn, David Minnen, Yong Cheng, Vighnesh Birodkar, Agrim Gupta, Xiuye Gu, Alexander~G. Hauptmann, Boqing Gong, Ming-Hsuan Yang, Irfan Essa, David~A. Ross, and Lu~Jiang.
\newblock Language model beats diffusion -- tokenizer is key to visual generation, 2024.
\newblock URL \url{https://arxiv.org/abs/2310.05737}.

\bibitem[Wang et~al.(2024{\natexlab{c}})Wang, Zheng, Ye, Xue, Huang, and Gu]{DPLM2}
Xinyou Wang, Zaixiang Zheng, Fei Ye, Dongyu Xue, Shujian Huang, and Quanquan Gu.
\newblock {DPLM}-2: {A} {Multimodal} {Diffusion} {Protein} {Language} {Model}, October 2024{\natexlab{c}}.
\newblock URL \url{http://arxiv.org/abs/2410.13782}.
\newblock arXiv:2410.13782 [cs].

\bibitem[Kuang et~al.(2025)Kuang, Liu, Wang, Sun, Ji, and Wu]{PDFBench}
Jiahao Kuang, Nuowei Liu, Jie Wang, Changzhi Sun, Tao Ji, and Yuanbin Wu.
\newblock {PDFBench}: {A} {Benchmark} for {De} novo {Protein} {Design} from {Function}, September 2025.
\newblock URL \url{http://arxiv.org/abs/2505.20346}.
\newblock arXiv:2505.20346 [cs].

\bibitem[Jing et~al.(2023)Jing, Erives, Pao-Huang, Corso, Berger, and Jaakkola]{Eigenfold}
Bowen Jing, Ezra Erives, Peter Pao-Huang, Gabriele Corso, Bonnie Berger, and Tommi Jaakkola.
\newblock Eigenfold: Generative protein structure prediction with diffusion models.
\newblock \emph{arXiv preprint arXiv:2304.02198}, 2023.

\bibitem[Campbell et~al.(2024)Campbell, Yim, Barzilay, Rainforth, and Jaakkola]{MultiFlow}
Andrew Campbell, Jason Yim, Regina Barzilay, Tom Rainforth, and Tommi Jaakkola.
\newblock Generative flows on discrete state-spaces: Enabling multimodal flows with applications to protein co-design.
\newblock \emph{arXiv preprint arXiv:2402.04997}, 2024.

\bibitem[Xiao et~al.(2026)Xiao, Lin, Chen, Wang, Chen, Li, Sun, Dai, Wu, and Ye]{Stella}
Hongwang Xiao, Wenjun Lin, Xi~Chen, Hui Wang, Kai Chen, Jiashan Li, Yuancheng Sun, Sicheng Dai, Boya Wu, and Qiwei Ye.
\newblock Stella: A multimodal llm for protein functional annotation via unified sequence-structure encoding.
\newblock In \emph{Findings of the Association for Computational Linguistics: ACL 2026}, pages 25030--25049, 2026.

\bibitem[Hsieh et~al.(2025)Hsieh, Wang, Zhang, Xue, Ye, Huang, Zheng, and Gu]{DPLM2.1}
Cheng-Yen Hsieh, Xinyou Wang, Daiheng Zhang, Dongyu Xue, Fei Ye, Shujian Huang, Zaixiang Zheng, and Quanquan Gu.
\newblock Elucidating the design space of multimodal protein language models.
\newblock \emph{arXiv preprint arXiv:2504.11454}, 2025.

\end{thebibliography}
\bibliographystyle{unsrtnat}

\clearpage
\beginappendix
\FloatBarrier
\section{Diffusion Preliminaries}
\label{app:diffusion-preliminaries}

Table~\ref{tab:app-diffusion-notation} summarizes the notation used throughout this appendix.

\begin{table}[!ht]
\centering
\caption{Summary of notation.}
\label{tab:app-diffusion-notation}
\small
\renewcommand{\arraystretch}{1.08}
\begin{tabular}{@{}l@{\hspace{1em}}p{0.66\linewidth}@{}}
\toprule
\textbf{Symbol} & \textbf{Description} \tabularnewline
\midrule
$\mathcal{V},\ V$ & \raggedright Token vocabulary and its size (excluding the mask) \tabularnewline
$x_0,\ x_t$ & \raggedright Clean and corrupted sequences of length $L$ \tabularnewline
$x_0^{i},\ x_t^{i}$ & \raggedright Corresponding one-hot tokens at position $i$ \tabularnewline
$\mathbf{M}$ & \raggedright Absorbing mask token, $e_{V+1}$ \tabularnewline
\addlinespace[3pt]
$s,\ t$ & \raggedright Diffusion times, $0\leq s<t\leq1$ \tabularnewline
$\alpha_t$ & \raggedright Token-retention probability, $\alpha_t=1-t$ \tabularnewline
$Q_t$ & \raggedright Forward transition matrix from time $0$ to $t$ \tabularnewline
$q(\cdot\mid\cdot)$ & \raggedright Forward or posterior probability mass function \tabularnewline
$\operatorname{Cat}(z;\pi)$ & \raggedright Categorical probability mass at $z$ with vector $\pi$ \tabularnewline
\addlinespace[3pt]
$\pi_\theta^{i}(x_t)$ & \raggedright Predicted clean-token probability vector \tabularnewline
$p_\theta(\cdot\mid\cdot)$ & \raggedright Model probability mass function \tabularnewline
$p_{\text{data}}$ & \raggedright Clean-data distribution \tabularnewline
$\mathbf{1}\!\left[\cdot\right]$ & \raggedright Indicator function; $\mathbf{1}$ alone is the all-ones vector \tabularnewline
\addlinespace[3pt]
$b,\ K$ & \raggedright Block size and number of blocks, $K=\lceil L/b\rceil$ \tabularnewline
$x_0^{[k]},\ x_0^{[<k]}$ & \raggedright Current block and preceding blocks \tabularnewline
$c$ & \raggedright Observed conditioning context \tabularnewline
\bottomrule
\end{tabular}
\end{table}

\subsection{Forward Transition Process}
\label{app:forward-diffusion}

Let $x_0=(x_0^1,\ldots,x_0^L)$ be a clean sequence sampled from $p_{\text{data}}(x_0)$. Each token
is represented by a one-hot column vector in $\{e_1,\ldots,e_V\}\subset\mathbb{R}^{V+1}$, and
$\mathbf{M}=e_{V+1}$ denotes the absorbing mask state. We write $\operatorname{Cat}(z;\pi)$ for the
categorical probability mass evaluated at $z$ with probability vector $\pi$; thus
$q(\cdot\mid\cdot)$ denotes a probability mass function, not a probability vector.

The forward process is defined by the cumulative transition matrix
\begin{equation}
\begin{aligned}
Q_t
&=\alpha_t I+(1-\alpha_t)\mathbf{1}\mathbf{M}^{\top}
\in\mathbb{R}^{(V+1)\times(V+1)},
\end{aligned}
\label{eq:app-transition}
\end{equation}
where $t\in[0,1]$, $\alpha_t$ is a monotonically decreasing retention schedule with $\alpha_0=1$ and
$\alpha_1=0$, and $\mathbf{1}$ is the all-ones column vector. Rows index the initial state and
columns index the resulting state. A token retains its identity with probability $\alpha_t$ and
becomes masked with probability $1-\alpha_t$; the mask state is absorbing.

Applying this transition independently at each position gives
\begin{equation}
\begin{aligned}
q(x_t\mid x_0)
&=\prod_{i=1}^{L}q(x_t^{i}\mid x_0^{i}),\\
q(x_t^{i}\mid x_0^{i})
&=\operatorname{Cat}\!\left(x_t^{i};Q_t^{\top}x_0^{i}\right)\\
&=\operatorname{Cat}\!\left(
 x_t^{i};\alpha_t x_0^{i}+(1-\alpha_t)\mathbf{M}
\right).
\end{aligned}
\label{eq:app-forward}
\end{equation}
For the linear schedule $\alpha_t=1-t$, each position is masked with probability $t$, and
$x_1=(\mathbf{M},\ldots,\mathbf{M})$.

\subsection{Reverse Denoising Process}
\label{app:reverse-diffusion}

For $0\leq s<t\leq1$, the forward posterior is
\begin{equation}
\begin{aligned}
q(x_s^{i}\mid x_t^{i},x_0^{i})
&=\begin{cases}
\operatorname{Cat}\!\left(x_s^{i};x_0^{i}\right),
&x_t^{i}\neq\mathbf{M},\\[5pt]
\operatorname{Cat}\!\left(
 x_s^{i};
 \dfrac{\alpha_s-\alpha_t}{1-\alpha_t}x_0^{i}
 +\dfrac{1-\alpha_s}{1-\alpha_t}\mathbf{M}
\right),
&x_t^{i}=\mathbf{M}.
\end{cases}
\end{aligned}
\label{eq:app-posterior}
\end{equation}
Let $\pi_\theta^{i}(x_t)\in\mathbb{R}^{V+1}$ denote the predicted clean-token probability vector at
position $i$, with zero mass on $\mathbf{M}$. Its corresponding probability mass function is
\begin{equation}
\begin{aligned}
p_\theta(x_0^{i}\mid x_t)
&=\operatorname{Cat}\!\left(x_0^{i};\pi_\theta^{i}(x_t)\right).
\end{aligned}
\label{eq:app-clean-prediction}
\end{equation}
Substituting this vector for the unknown clean token in the masked branch gives, under
$\alpha_t=1-t$,
\begin{equation}
\begin{aligned}
p_\theta(x_s^{i}\mid x_t)
&=\begin{cases}
\operatorname{Cat}\!\left(x_s^{i};x_t^{i}\right),
&x_t^{i}\neq\mathbf{M},\\[5pt]
\operatorname{Cat}\!\left(
 x_s^{i};\dfrac{t-s}{t}\pi_\theta^{i}(x_t)
 +\dfrac{s}{t}\mathbf{M}
\right),
&x_t^{i}=\mathbf{M}.
\end{cases}
\end{aligned}
\label{eq:app-reverse}
\end{equation}
This ancestral update preserves visible tokens and progressively reveals masked tokens as time
decreases from $1$ to $0$.

\subsection{Masked Diffusion Language Modeling Objective}
\label{app:denoising}

Under the substitution parameterization of the absorbing-state, the negative evidence lower bound in
continuous-time reduces to a weighted reconstruction loss over masked positions~\citep{MDLM}. For a
general retention schedule, the positive weight on negative log-probability is
$-\alpha'_t/(1-\alpha_t)$. Substituting $\alpha_t=1-t$ yields
\begin{equation}
\begin{aligned}
\mathcal{L}_{\text{MDLM}}(\theta)
&=
-\mathbb{E}_{t \sim \mathcal{U}(0, 1), x_0 \sim p_{\text{data}}(x_0)}
\mathbb{E}_{x_t \sim q(x_t \mid x_0)}
\left[
\frac{1}{t}
\sum_{i=1}^{L}
\mathbf{1}\!\left[x_t^{i}=\mathbf{M}\right]
\log p_\theta\!\left(x_0^{i}\mid x_t\right)
\right].
\end{aligned}
\label{eq:app-diffusion-loss}
\end{equation}
Only masked positions contribute to the loss. The predictor uses the same-position clean token as
its target, with bidirectional access to visible tokens. Time is implicit in the corruption; the
notation does not require explicit time conditioning. Numerical truncation near $t=0$ and token
normalization are specified with the training configuration.

\subsection{Block-Wise Diffusion}
\label{app:block-diffusion}

Partition the sequence into $K=\lceil L/b\rceil$ contiguous blocks of maximum size $b$, denoted by
$x_0^{[1]},\ldots,x_0^{[K]}$. Block diffusion factorizes the sequence distribution as
\begin{equation}
\begin{aligned}
p_\theta(x_0)
&=\prod_{k=1}^{K}p_\theta\!\left(x_0^{[k]}\mid x_0^{[<k]}\right).
\end{aligned}
\label{eq:app-block-factorization}
\end{equation}
Each conditional is modeled by the same masking and denoising construction, with the preceding
blocks supplied as fixed context. Tokens within the current block interact bidirectionally, while
future blocks remain inaccessible. Training uses clean preceding blocks; inference uses completed
generated blocks. This combines autoregressive ordering across blocks with diffusion within
blocks~\citep{BD3LM}.

\subsection{Conditional Generation Across Modalities}
\label{app:unified-conditioning}

For a task with observed context $c$, replace $p_\theta(x_0^{i}\mid x_t)$ with $p_\theta(x_0^{i}\mid
x_t,c)$ and draw $(x_0,c)$ from the paired data distribution. The context contains the supplied
modality and task instruction, while $x_0$ denotes the target modality. For folding, the context
contains an amino-acid sequence and the target contains backbone tokens; inverse folding exchanges
these roles. Text understanding and text-conditioned sequence generation follow the same
construction. In conditional answer-only training, the context stays uncorrupted and only target
positions enter the masking process and loss. Full-sequence supervision is a separate training
configuration.

\FloatBarrier
\section{SFT Dataset Details}
\label{app:sft-dataset-details}

This appendix presents the task templates and corpus used for supervised
fine-tuning (SFT).

\subsection{Task Templates}
\label{app:task-examples}

Table~\ref{tab:app-task-templates} shows examples of the six directed tasks
over sequences ($S$), backbones ($B$), and text ($T$).

\begin{table}[!htbp]
\centering
\caption{SFT examples and data sources. Italicized placeholders replace
biological token spans; \{\textit{motif}\} denotes a segment within the generated
sequence. Modality and motif special tokens are omitted for clarity.}
\label{tab:app-task-templates}
\small
\begingroup
\newcommand{\appfield}[1]{\{\textit{#1}\}}
\renewcommand{\arraystretch}{1.12}
\begin{tabular}{@{}
p{0.07\linewidth}@{\hspace{0.015\linewidth}}
p{0.11\linewidth}@{\hspace{0.015\linewidth}}
p{0.40\linewidth}@{\hspace{0.015\linewidth}}
p{0.06\linewidth}@{\hspace{0.015\linewidth}}
p{0.28\linewidth}@{}}
\toprule
\textbf{Task} & \textbf{Source} & \textbf{Prompt} & \textbf{CoT} & \textbf{Answer} \tabularnewline
\midrule
S2B & \raggedright DPLM2;\newline UC40
& \raggedright Generate a plausible backbone structure for the sequence below: \appfield{sequence}
& \centering $\times$
& \raggedright \appfield{backbone} \tabularnewline
\addlinespace[6pt]
B2S & \raggedright DPLM2;\newline UC40
& \raggedright \appfield{backbone} Use the protein structure above to construct a plausible sequence.
& \centering $\times$
& \raggedright \appfield{sequence} \tabularnewline
\addlinespace[6pt]
S2T & \raggedright Swiss-Prot
& \raggedright Examine the protein sequence and indicate which targeting signals (e.g., signal peptide, mitochondrial targeting peptide, nuclear localization signal) are present. The protein is \appfield{sequence}
& \centering $\checkmark$
& \raggedright The protein contains an N-terminal signal peptide; no mitochondrial targeting sequence or nuclear localization signal is predicted. \tabularnewline
\addlinespace[6pt]
T2S & \raggedright Swiss-Prot
& \raggedright Formulate an original protein sequence featuring the Photosystem II PsbT protein family. Design a complete, biologically relevant protein that exhibits the core features and actions of this protein family. Embrace the normal structural conformations, molecular dimensions, and biological activities of the Photosystem II PsbT (protein family).
& \centering $\checkmark$
& \raggedright The \appfield{motif} is a single N-terminal hydrophobic helix followed by a short basic C-terminal tail; it could be constructed into the \appfield{sequence}. \tabularnewline
\addlinespace[6pt]
B2T & \raggedright OPI-Struc
& \raggedright \appfield{backbone} Could you share a detailed analysis of the function attributed to the protein?
& \centering $\times$
& \raggedright Involved in oxygen transport from the lung to the various peripheral tissues. \tabularnewline
\addlinespace[6pt]
T2B & \raggedright Prot2Text
& \raggedright Generate a protein backbone of approximately 354 residues for a Tryptophanyl-tRNA synthetase with the following function: Catalyzes the attachment of tryptophan to tRNA(Trp).
& \centering $\times$
& \raggedright \appfield{backbone} \tabularnewline
\bottomrule
\end{tabular}
\endgroup
\end{table}

All examples use a shared user--assistant format, with role delimiters omitted.
CoT marks whether the source instance includes reasoning; Answer shows the
final response after that reasoning span.

\FloatBarrier
\subsection{Dataset Statistics}
\label{app:dataset-statistics}

\begin{figure*}[!ht]
\centering
\includegraphics[width=0.8\linewidth]{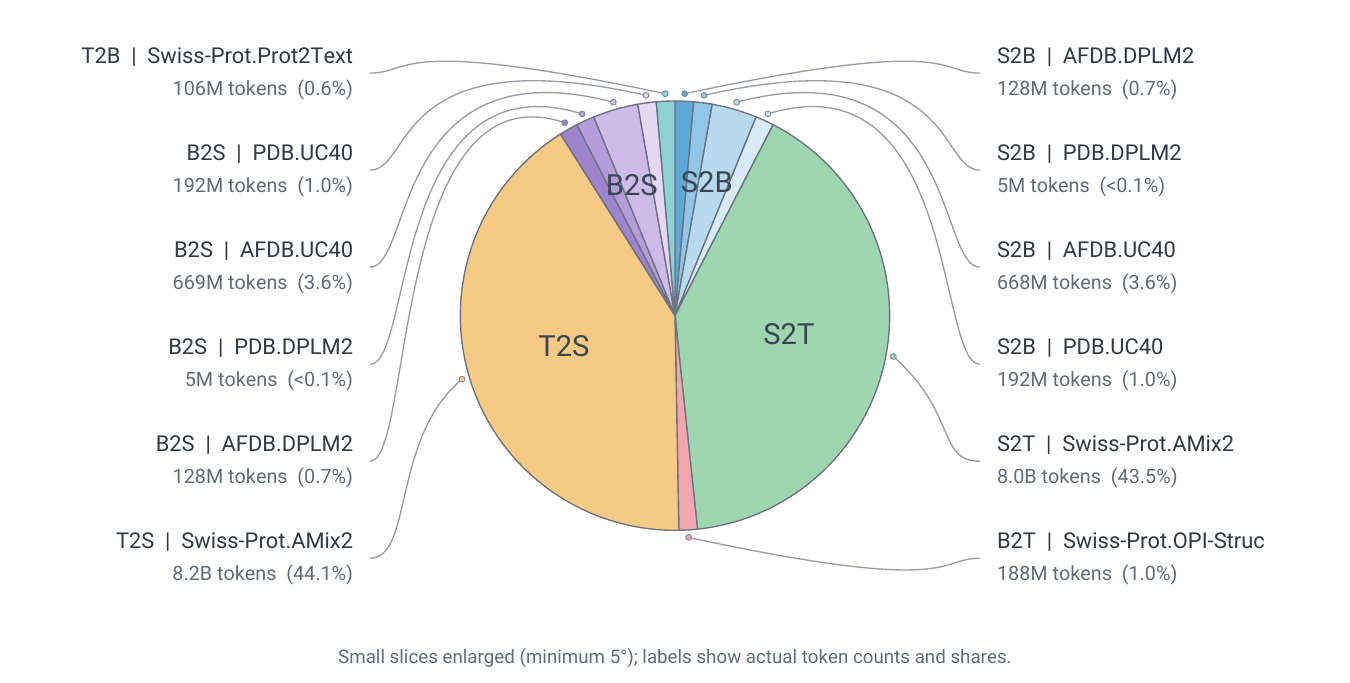}
\caption{SFT corpus by task and source (\textit{database.collection}).
Labels show actual token counts and shares (B: billion; M: million).
Small slices are enlarged to at least $5^{\circ}$ for readability, so areas
are not proportional to token counts.}
\label{fig:dataset-token-statistics}
\end{figure*}

Figure~\ref{fig:dataset-token-statistics} summarizes the $18.5$B-token SFT
corpus before further dataset sampling during training. Corpus proportions
therefore differ from the effective training mixture.

\FloatBarrier
\section{Benchmark Evaluation Protocol}
\label{app:benchmark-evaluation-protocol}

\paragraph{Folding.}
We evaluate 632 monomeric targets: 183 from CAMEO 2022 and 449 from the PDB date split. Predictions
and native structures are mapped to the reference sequence, with native extensions cropped and
missing atoms masked. We report TMca, TMbb, and RMSDbb, averaged per protein within each split. TMca
uses C$\alpha$ coordinates, whereas TMbb uses backbone N/C$\alpha$/C coordinates; both are computed
with TM-align and normalized by the native residue count after filtering to paired-valid residues.
RMSDbb is the root mean squared distance over paired-valid backbone atoms after a single Kabsch
alignment, rather than TM-align's local RMSD. Higher TM scores and lower RMSDbb indicate better
predictions. Paired table entries show mean TMbb/TMca.

\paragraph{Inverse folding.}
We use the same 183 CAMEO 2022 and 449 PDB targets and report amino-acid recovery (AAR) and
self-consistency TMbb (scTMbb). For a generated sequence $\hat{s}$ and reference $s$ of equal length
$L$, AAR is $100L^{-1}\sum_{j=1}^{L}\mathbf{1}[\hat{s}_j=s_j]$, without post-hoc sequence
realignment. Each generated sequence is folded once with ESMFold v1; scTMbb compares this predicted
structure with the native backbone using the same atom masks and native-normalized TMbb protocol as
folding. Both metrics are averaged per protein, with higher values indicating better performance.
Unavailable folding scores are excluded from scTMbb aggregation while retaining AAR and recording
metric coverage.

\paragraph{Sequence understanding.}
GeneralQA contains 481 questions: Function \& Family (150), Interaction \& Binding (152), Location
\& Modification (153), and Other (26; 19 physicochemical-property and seven structural-composition
questions). A Gemini-based judge compares each generated response with the reference answer for
biological equivalence, accepting equivalent concise descriptions and rejecting substantive
inconsistencies or ambiguity. Each question receives a binary correct/incorrect decision. Accuracy
is $100N_{\mathrm{correct}}/N$ within each category; overall accuracy pools all 481 questions rather
than averaging category percentages. Judge confidence does not weight the score.

\paragraph{Structure understanding.}
StructureMCQA comprises 3,918 four-option questions from OPI-Struc. We extract the predicted option
from the final answer and report exact-choice accuracy, $100N_{\mathrm{correct}}/3918$; unrecognized
or conflicting explicitly marked choices count as incorrect. The BioMatrix entry uses a separately
marked nearest-option protocol because its outputs often do not follow the requested format: pure
option letters are mapped directly, while free-text responses are mapped to the semantically closest
option by a Gemini-based judge without access to the gold answer, then compared with the reference
locally. This differs from direct-choice extraction; BioMatrix also receives sequence and structure
inputs, whereas Trion-1 receives backbone inputs.

\paragraph{Sequence design.}
We evaluate 870 InterPro-based prompts: Family (569), Superfamily (57), Domain (239), and Site (5),
preserving repeated benchmark records. Success rate (SR) is the percentage of all 870 prompts
yielding an accepted sequence for evaluation, not functional correctness. Quality metrics are
averaged over finite per-sequence scores and displayed on a 0--100 scale. Repetition is
$\mathrm{Rep}_n=1-|U_n|/(L-n+1)$ for $n\in\{2,5\}$, where $U_n$ is the set of distinct overlapping
$n$-grams in a sequence of length $L$; it is zero when $L\leq n$, and lower is better. Foldability
is measured by ESMFold v1 pLDDT, averaged over atoms in the predicted structure; the PDFBench
implementation folds only the first 1,024 residues of longer sequences. Functional recovery is
$\mathrm{IPRRecovery}=|R\cap G|/|R|$, where $R$ and $G$ are the reference and generated sequences'
InterPro annotation sets; an empty reference set yields an unavailable score. Hard sequence novelty
is one minus the maximum MMseqs2 alignment identity among retrieved UniProt hits, or one when a
successful search returns no hits. Higher pLDDT, IPRRecovery, and novelty are better. The table uses
the first stored candidate per prompt, with the marked likelihood-ranked best-of-ten exception for
Pinal; failed parses remain in the SR denominator, and metric-specific coverage is retained.

\end{document}